\documentclass[11pt,a4paper]{article}

\usepackage[utf8]{inputenc}
\usepackage[T1]{fontenc}
\usepackage{lmodern}
\usepackage{geometry}
\usepackage{graphicx}
\usepackage{amsmath,amssymb,amsfonts}
\usepackage{authblk} %
\usepackage{float}
\usepackage{multirow}
\usepackage[table]{xcolor}
\usepackage{array}
\usepackage{placeins}
\usepackage{microtype}
\usepackage{hyperref}
\usepackage{booktabs}
\usepackage{cite}
\usepackage[caption=false]{subfig}
\ifdefined\pdfminorversion\pdfminorversion=7\fi

\hypersetup{
    colorlinks=true,
    linkcolor=blue,
    filecolor=magenta,      
    urlcolor=cyan,
    citecolor=blue
}

\title{From WSIs to Bulk, Spatial and Single-Cell Transcriptomics via Multi-Scale Graph Neural Networks}

\author[1]{Miguel Andrade}
\author[2]{José Almeida}
\author[1]{Carlos Santiago}
\author[1]{Catarina Barata}

\affil[1]{Departamento de Engenharia Electrotécnica
e de Computadores, Instituto Superior Técnico,
Lisboa, Portugal}

\affil[2]{Champalimaud Foundation, Lisbon, Portugal}

\date{}

\begin{document}
\maketitle
\begin{abstract}
Understanding the spatial distribution of gene expression is key to studying complex diseases like cancer. While Spatial Transcriptomics (ST) and Single-Cell Transcriptomics (scT) provide high-resolution insights, they remain expensive and scarce. In contrast, standard histopathology Whole-Slide Images (WSIs) paired with bulk RNA sequencing are widely available. This work introduces a Multi-Scale Graph Neural Network (GNN) capable of simultaneously predicting bulk, spatial, and single-cell transcriptomics directly from WSIs. By extracting morphological features using a Pathology Foundation Model (PFM) (UNI2-h) and constructing spatial graphs at both patch and cell levels, the model learns the hierarchical relationships within the tissue. The network is trained on a combined dataset of The Cancer Genome Atlas Program (TCGA), Visium, and Xenium samples across lung, kidney, and breast tissues. A hybrid training strategy is employed, leveraging direct supervision alongside weakly supervised consistency losses to ensure predictions align across biological scales. Results show the model accurately predicts gene expression at the slide level and moderately at the patch level (strongest on Xenium), and achieves moderate correlation at the challenging single-cell level. Furthermore, downstream tasks explore the clinical relevance of these predictions, demonstrating their utility in survival analysis, spatial hazard mapping, and tumor classification via molecular signatures.
\end{abstract}

\section{Introduction}
Understanding genes is key to grasping how our body works and evolves, as well as making sense of complex diseases such as cancer. That is why ST and scT may provide critical insights for understanding and effectively treating tumors. The rapid evolution of ST and scT has transformed our understanding of tissue biology by enabling the systematic measurement of expression levels for all or most genes throughout tissue space \cite{spatialtranscriptomicsintro}. Despite their revolutionary potential, these methods remain constrained by high experimental costs (thousands of dollars per slide), and are not yet routine in most hospitals. On the other hand, histopathology is already a common practice in cancer diagnosis, producing hematoxylin and eosin-stained WSIs. WSIs are high-resolution digital representations of histopathology slides that capture the complete morphological context of the tissue. They preserve cellular structures, tissue architecture and spatial organization of cells.

Although ST provides a good understanding of the spatial distribution of genes, it does not offer the finest granularity available. In one patch there are often dozens to hundreds of cells, so the expression of each spot is actually the mixture of the gene expressions of each cell within each patch. scT achieves the finest granularity available, providing the gene expression for each detected cell in a WSI. Additionally, WSIs are often used as a conclusive diagnostic tool because many diseases, including cancer, are characterized by abnormal spatial organization within tissues \cite{Damjanov:1996:Histopathology}. By mapping mRNA within tissues, we improve our ability to visualize and interpret biological processes, since abnormal mRNA expression is frequently associated with human tumor initiation, development, progression and metastasis \cite{LiWang2021_bulk_sc_spatialRNAseq}. 

The relationship between tissue morphology and gene expression arises from the fact that tissue architecture and histological appearance are a direct manifestation of underlying genes at play. Differences in cell type, microenvironmental context and the presence of a tumor can result in characteristic morphological signatures. These morphological features can share a high amount of information with gene expression \cite{Chelebian2025_combining}. Therefore, using WSIs to predict ST and scT represents the next logical step, that not only assists doctors in diagnosing tumor slides but also enables a wide range of downstream tasks such as biomarker discovery and survival prediction.

ST and scT have advanced rapidly in both experimental and computational methods, yet there remains a critical limitation in accurately predicting transcriptomic profiles. Moreover, current deep learning models can infer partial gene expression patterns from histopathology slides. However, their predictions often show limited gene coverage, weak generalization across different tissues, and low accuracy in reconstructing the full transcriptome. While the acquisition of spatial data is expensive and time-consuming, bulk gene expression data, which provides the average gene expression across the whole tissue, is far more accessible and abundant. However, the translation from bulk-level to spatial gene expression profiles remains under-explored. 

To bridge this gap, this work leverages three types of data: (1) WSIs paired with bulk gene expression; (2) WSIs paired with ST; and (3) WSIs paired with scT. WSIs are typically divided into smaller, fixed-size patches to enable efficient processing by deep learning models. ST requires that patches align with the spots where mRNA was sequenced (arranged across the WSI in a predefined grid pattern to enable spatially resolved transcriptomic profiling). In addition, in cases where scT data is available, each detected cell has an associated gene expression profile, although these profiles can be very sparse, since each cell may have counts for only a few of the genes in the panel. On the other hand, if only bulk gene expression is available, the spatial information is lost and the gene expression corresponds to the mixture of the whole tissue. This poses a challenge, as the bulk gene expression does not correspond exactly to the WSI and the spatial relationship between the WSI and the genomic data is unknown \cite{Cooper2018PanCancerPathology}. Despite this, the bulk gene expression can be reasonably approximated as representative of the average gene expression across the WSI, given that the WSI and the sampled tissue originate from the same primary tumor.

The results obtained by recent studies \cite{huang2025stflow,zou2025carhe,dashtiahangar2025multiomics,mosby} support the hypothesis that, in most cases, morphology encodes sufficient information to approximate gene expression patterns. Consequently, deep learning models trained on paired data WSI-transcriptomics can exploit these correlations to reconstruct spatial gene profile maps directly from WSIs.

Most existing models that have the capability of predicting at the single-cell level employ the same training strategy: treating each spot as a bag of cells and using Multiple Instance Learning (MIL) to predict the spot's gene expression. This way, the model learns how the aggregation of the cell's gene expression should look like through weakly supervised learning. This strategy was employed in a state of the art model, DeepSpot2Cell \cite{nonchev2025deepspot2cell}, which leverages a PFM and spatial multi-level context to predict scT using spot-level supervision. A similar approach was proposed in \cite{chadoutaud2026scellst}. The methodology is the same, as the spot is represented as a bag of cells, and the cells' gene expression is learned through MIL.

In this work, a Multi-Scale GNN was developed to predict gene expression from WSIs at three levels: bulk, spatial and single-cell. The consistency losses encourage the predictions at these levels to agree.

\section{Results}

\subsection{Overall pipeline architecture}
\label{section:overview}

\begin{figure*}[!t]
    \centering
    \includegraphics[width=\textwidth]{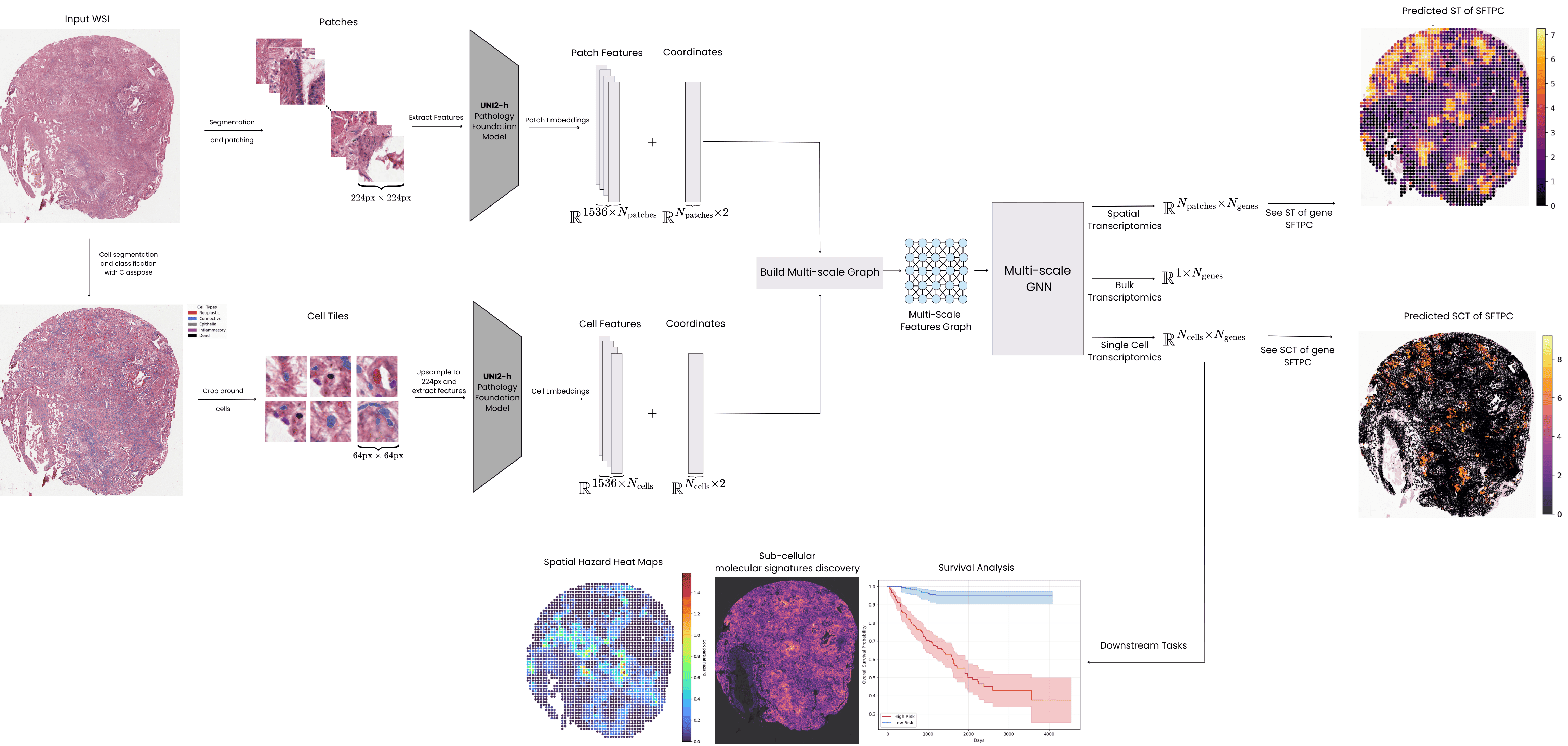}
    \caption{Proposed model's overall pipeline architecture.}
    \label{fig:overall_pipeline}
\end{figure*}

The overall pipeline can be seen in Figure~\ref{fig:overall_pipeline}, which shows how an unseen WSI would be processed and the inference pipeline of our model. By extracting morphological features using a PFM (UNI2-h) and constructing spatial graphs at both patch and cell levels, the model learns the hierarchical relationships within the tissue. Cell information is passed to the patch branch, and patch information is aggregated to obtain the slide-level prediction. The network is trained on a combined dataset of TCGA, Visium, and Xenium samples across lung, kidney, and breast tissues. A hybrid training strategy is employed, leveraging direct supervision alongside weakly supervised consistency losses to ensure predictions align across biological scales.

\subsection{Visual features correlate with gene expression}
\label{section:visual_features}

This study relies on a fundamental hypothesis: gene expression is predictable, to some degree, from the visual features extracted from WSIs.

Visium slides contain a high amount of batch effects both in the feature space and gene expression space. This creates a slide-level confound, where patches from the same slide cluster together in both spaces not necessarily due to shared biology, but due to these batch effects. This phenomenon has been documented for PFMs, which have been shown to encode strong site-specific signatures in their principal components \cite{komen2024histopathologicalfoundationmodelseliminate}.

The patch features, extracted using UNI2-h, were compressed to $100$ components using PCA. The $17{,}038$-dimensional gene expression vectors were compressed into $300$ principal components. To assess whether the correspondence between embeddings and transcriptomics reflects a biological signal, Harmony \cite{Korsunsky2019} was applied to both spaces to correct for \textit{slide\_id} as a batch covariate. The number of clusters was computed using the Leiden algorithm, with $22$ clusters in the patch feature space and $20$ clusters in the gene expression space.

The t-SNE plot of the gene expressions was then colored using the cluster assignment from the patch feature space. The goal was to investigate whether clusters in the patch embedding space correspond to clusters in the gene expression space. As shown in Supplementary Figure~\ref{fig:tsne_genes_features_clustered_harmonized}, this correspondence does indeed occur. The cluster purity was $55\%$, the Adjusted Rand Index (ARI) was $0.1251$, and the Normalized Mutual Information (NMI) was $0.2773$. While Harmony correction cannot guarantee the complete removal of every possible batch effect, the fact that a substantial correlation persists after explicit slide-level correction indicates that the observed relationship is not solely attributable to batch effects.
This is further supported by computing the PCC between gene expressions in the same cluster, which was found to be $0.3007$. While moderate, it suggests that patches with similar features tend to exhibit similar gene expression profiles. Furthermore, the PFM used (UNI2-h) is not ``gene-aware'', that is, it was only pre-trained on general histopathology images without exposure to ST data. This general-purpose encoder, which was pre-trained without exposure to ST data, recovers patterns that align with the gene expression space. Some of this alignment is expected, since cell types that are distinguishable under the microscope are themselves defined by the expression of specific genes, so the observed correlation likely reflects, at least in part, cell-type composition rather than finer morphological cues. Nonetheless, it indicates that the visual features capture information that is informative of the underlying transcriptomic state.

To further study this relationship, Classpose \cite{classpose} classifies each cell into one of the five available labels: neoplastic, connective, inflammatory, epithelial, and dead. In Supplementary Figure~\ref{fig:st_colored_w_cell_percentage}, it is clear that certain gene profiles are related to the percentage of the cell type present in the corresponding patch. For example, for neoplastic cells (tumor-related), there are only certain clusters that contain a high percentage of this cell, meaning certain gene expression profiles are associated with this cell type.

\subsection{Gene expression prediction at bulk, spatial and single-cell levels}
\label{section:quantitative_results}

The results are organized by dataset (TCGA, Visium or Xenium) and by tissue type (lung, kidney, breast and overall cohort). This enables a more detailed analysis, uncovering the performance of each cohort in the dataset. The metrics are reported as the mean and standard deviations across five cross-validation folds. Table~\ref{tab:prediction_summary} presents the main performance metrics obtained by the proposed model across datasets. The bulk, patch and cell PCC values in this table compare expression across genes within a sample. The gene-wise mean PCC curves in Figure~\ref{fig:results_panel}A and B compare expression across samples for a given gene and report the mean for the selected top-predictive genes. The patch and cell AUROC values are computed per gene within each slide: spots (or cells) whose true expression of the gene is in the top quartile are labelled as high, those in the bottom quartile as low, the middle $50\%$ are discarded, and the AUROC measures how well the predicted values separate the two groups. The AUROC is then averaged over genes and over the test slides of each fold (see Section~\ref{methods:metrics}).

\begin{table}[tbp]
\centering
\caption{Proposed model's metrics across datasets.
Metrics are reported as the mean and standard deviations
across five cross-validation folds.}
\label{tab:prediction_summary}
\vspace{5pt}
\footnotesize
\setlength{\tabcolsep}{3pt}
\begin{tabular}{lccc}
\toprule
Metric
& \shortstack{TCGA\\$N=1724$}
& \shortstack{Visium\\$N=105$}
& \shortstack{Xenium\\$N=47$} \\
\midrule
Bulk PCC
& $0.881\pm0.010$
& $0.774\pm0.146$
& $0.797\pm0.057$ \\

Patch PCC
& --
& $0.509\pm0.026$
& $0.786\pm0.041$ \\

Cell PCC
& --
& --
& $0.396\pm0.021$ \\

Patch AUROC
& --
& $0.554\pm0.004$
& $0.703\pm0.011$ \\

Cell AUROC
& --
& --
& $0.649\pm0.011$ \\
\bottomrule
\end{tabular}
\end{table}

Analyzing the Visium results in Supplementary Table~\ref{tab:visium_metrics}, a very distinct performance gap emerges between bulk-level and patch-level predictions. The model achieves a strong bulk PCC (overall: $0.774 \pm 0.146$), but the patch-level predictions drop to $0.509 \pm 0.026$, which shows a moderate linear relationship between the predictions and the ground-truth. This drop in performance is expected, since it reflects the inherent challenge of predicting ST data. Furthermore, lung samples exhibit the highest patch-level AUROC, although the spot-wise patch PCC is lower, the patch AUROC is the highest among all organs ($0.586$). Figure~\ref{fig:results_panel}A (chart A) corroborates this: lung exhibits the highest gene-wise PCC. This demonstrates that the model is able to capture distinct spatial profiles, consistently predicting higher expression values in regions that genuinely correspond to the top quartile of gene expression. It is worth noting that breast does not have a valid FDR value, this outcome is attributable to the small cohort size ($N = 7$), where each test split contains at most two breast samples, severely limiting statistical power in hypothesis testing.
\par
In the case of the Xenium dataset, the model maintains a strong predictive power at both the slide level (overall: $r=0.797$) and the patch level ($r=0.786$). On the other hand, the cell-level PCC is smaller at $0.396$, nonetheless this result still shows that the model is able to predict at single-cell level with a moderate positive linear relationship with the ground-truth values. Stratifying by cell type, the model exhibits a higher correlation in epithelial ($r=0.406$) and neoplastic ($r=0.456$) cells compared to dead ($r=0.330$), inflammatory ($r=0.340$) and connective ($r=0.367$) cells. Single-cell predictions could only be directly evaluated for the genes measured by Xenium.

Finally, for the TCGA cohort, the model shows high reliability in predicting the bulk gene expression across all tissue types. This is confirmed by the very strong correlations (overall: $r=0.881$) in Supplementary Table~\ref{tab:tcga_metrics}, along with the high gene-wise correlations in Figure~\ref{fig:results_panel} (A) (chart C). The fraction of significantly predicted bulk genes is also high ($72.6\% \pm 16.0\%$), confirming the model's capacity to predict global slide expression.

Figure~\ref{fig:results_panel}B compares the proposed model with DeepSpot2Cell and STFlow. DeepSpot2Cell achieves higher spot-level gene-wise PCC values than the proposed model. On Visium the difference is within the fold variability, while on Xenium it is larger than the fold variability. DeepSpot2Cell also attains a higher patch AUROC on both datasets (Visium $0.590$ vs $0.554$; Xenium $0.764$ vs $0.703$) (Supplementary Tables~\ref{tab:visium_metrics_different_models} and~\ref{tab:xenium_metrics_different_models}). Conversely, at cell level, the proposed model is clearly better than DeepSpot2Cell (cell PCC $0.396$ vs $0.349$; cell AUROC $0.649$ vs $0.523$). Differences in the supervision available to each model limit these comparisons: the proposed model has direct supervision at cell level, while DeepSpot2Cell uses a weakly supervised loss function. STFlow has lower gene-wise PCC in the displayed comparisons, although on Visium its patch-level metrics are comparable to the proposed model's.
\subsection{Spatial patterns in Visium and Xenium}
\label{section:visual_results}

To complement the quantitative metrics reported above, this section presents qualitative visual comparisons between predicted and ground-truth gene expression for individual slides from the Visium and Xenium test sets. For each slide, the displayed gene was selected as the one with the highest spatial expression variance within that specific slide, as high-variance genes tend to exhibit the clearest spatial patterns and therefore best illustrate whether the model is capturing meaningful morphological structure. 
\par
\begin{figure*}[tp]
\centering
\includegraphics[width=0.82\textwidth]{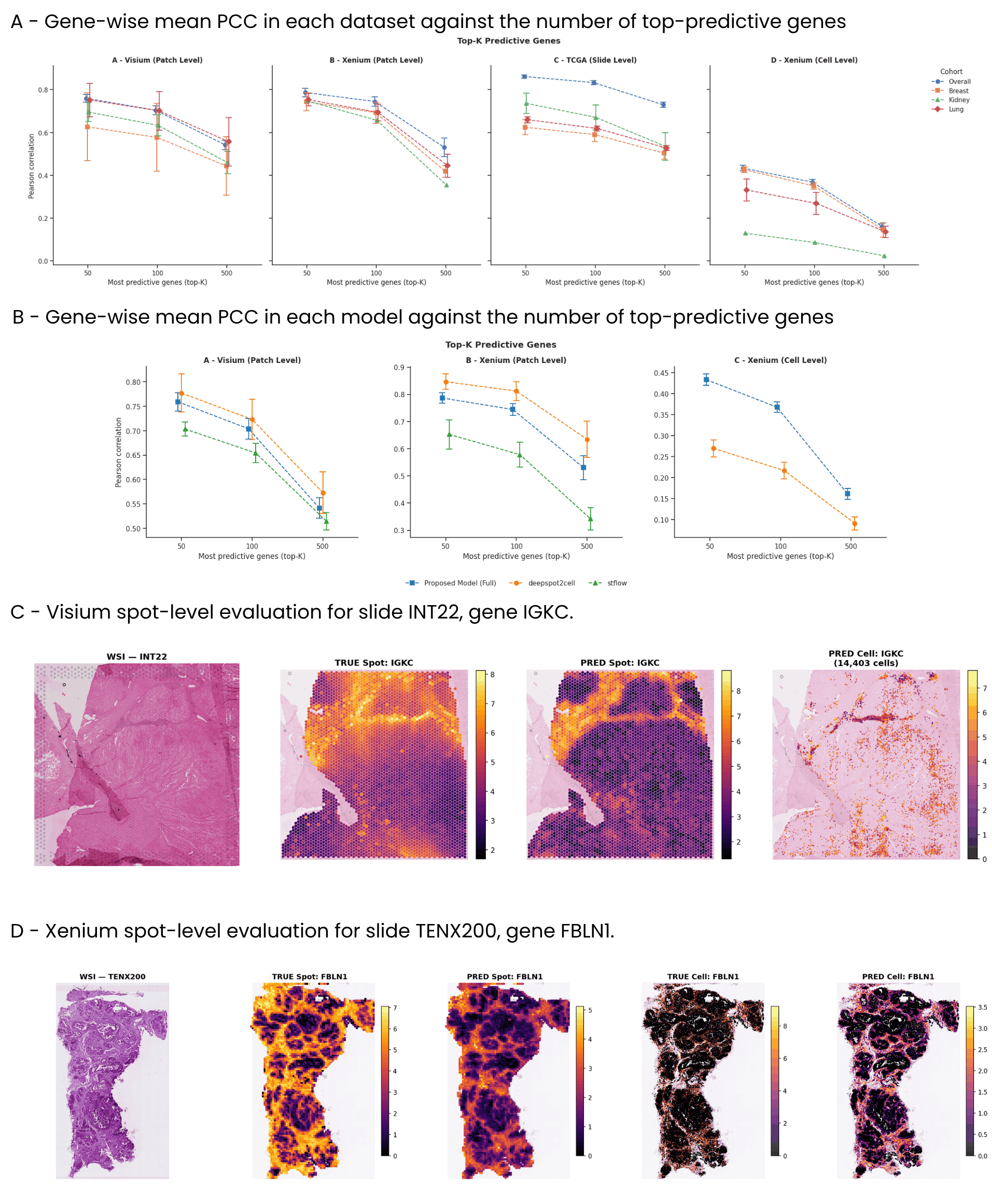}
\caption{\textbf{Multi-Scale GNN results.} \textbf{A}) Gene-wise mean PCC in each dataset against the number of top-predictive genes in the proposed model. \textbf{B}) Gene-wise mean PCC in each model against the number of top-predictive genes. \textbf{C}) Visium spot-level evaluation for slide INT22 (kidney), gene \textit{IGKC}. Left to right: WSI, true spot expression, predicted spot expression, predicted cell expression. \textbf{D}) Xenium spot-level evaluation for slide TENX200 (breast), gene \textit{FBLN1}. Left to right: WSI, true spot expression, predicted spot expression, true cell expression, predicted cell expression.}
\label{fig:results_panel}
\end{figure*}

Figure~\ref{fig:results_panel} (C) shows a Visium spot-level evaluation for slide INT22, with \textit{IGKC} as the selected gene. The predicted spatial expression map closely reproduces the main structural pattern of the ground truth.
\par
Figure~\ref{fig:results_panel} (D) presents the corresponding evaluation on the Xenium slide TENX200, at spot and single-cell resolution, for the gene \textit{FBLN1}. The predicted spot-level map visually reproduces the tumor morphology visible in the WSI, matching the true expression pattern with a spot-level PCC of $r=0.775$. At the single-cell level, the predicted expression is visibly noisier and the correlation drops accordingly ($r=0.458$), which is expected given the substantially higher spatial resolution and consequent sparsity of single-cell transcript counts.

\subsection{Predicted bulk expression supports survival analysis}
\label{subsection:survival_analysis}

In order to validate the quality of the bulk gene expression predictions made by the model, a survival analysis task was performed on both the prediction and the ground truth bulk gene expression, and the results were compared. Clinical outcome variables were obtained from the TCGA Pan-Cancer Clinical Data Resource (TCGA-CDR), a curated and standardized clinical annotation dataset for TCGA cohorts \cite{LIU2018400}. More specifically, for TCGA-KIRC, TCGA-LUAD and TCGA-BRCA the Disease-Specific Survival (DSS) time was extracted for each patient. For each cohort, three Cox models \cite{coxmodel} were trained: one on the $50$ PCs (obtained through PCA on the $17{,}038$ dimensional gene expression) of the true bulk gene expression, another on the $50$ PCs of the predicted bulk gene expression and the last on the $50$ PCs of the visual features embedding. The Cox models were trained and tested on all patients from each TCGA cohort using five-fold cross validation. Table~\ref{tab:survival_analysis_c_index} compares the C-index obtained using ground-truth bulk expression, predicted bulk expression, and patch visual features (mean-pooled UNI2-h embeddings) for each patient. Using the predicted instead of the ground-truth bulk expression lowers the C-index in all three cohorts, by $0.047$ in TCGA-LUAD, $0.083$ in TCGA-KIRC, and $0.029$ in TCGA-BRCA. The results also reveal that the gene expression is not a good predictor for the patient outcome in the TCGA-LUAD cohort, since the C-index is close to $0.5$, which means that the Cox model's risk scores have no meaningful relationship with the survival times. In contrast, for TCGA-KIRC and TCGA-BRCA the scores reveal an acceptable model, with C-index scores above $0.6$, across the three input types.

\begin{table}[h]
\centering
\caption{Comparison of Bulk C-index metrics across cohorts.}
\label{tab:survival_analysis_c_index}
\vspace{5pt}
\footnotesize
\setlength{\tabcolsep}{3pt}
\begin{tabular}{lccc}
\toprule
C-index
& \shortstack{TCGA-LUAD\\$N=393$}
& \shortstack{TCGA-KIRC\\$N=413$}
& \shortstack{TCGA-BRCA\\$N=834$} \\
\midrule
Real bulk      & $0.544$ & $0.748$ & $0.639$ \\
Predicted bulk & $0.497$ & $0.665$ & $0.610$ \\
Visual features & $0.555$ & $0.688$ & $0.730$ \\
\bottomrule
\end{tabular}
\end{table}

\subsection{Molecular signatures and spatial hazard heat maps}
\label{subsection:pathway_discovery}

The analysis made in the previous Section is important, but it has a limitation: since the $17{,}038$ genes were compressed into $50$ principal components, the contribution of individual genes to the patient's outcome can only be assessed indirectly. Each principal component is a weighted combination of all genes, so the genes that drive it can be identified from their weights in that component (also known as loadings). The Cox model coefficients can then be mapped back to the genes. However, every gene contributes to every component, and each component often mixes several biological processes, which makes it difficult to attribute an increase or decrease in risk to specific genes or pathways. To tackle this, Cox models were trained directly on the mean expression of established molecular signatures. A molecular signature or Hallmark gene set is a collection of genes that are associated to certain biological processes. 
\par
The gene sets used were extracted from the Molecular Signatures Database (MSigDB) \cite{LIBERZON2015417}, and the ones analyzed are: (1) proliferation, this molecular signature contains $200$ genes that are required for cellular division and cell cycle progression. Tumors over-express this signature to enable uncontrolled proliferation and rapid expansion of the tumor; (2) apoptosis, this gene set, with $161$ genes, is responsible for programmed cell death which eliminates damaged or abnormal cells. Tumor cells frequently evolve to suppress this pathway, enabling them to evade self-destruction; (3) angiogenesis, this hallmark, containing $36$ genes, represents the biological process of generating new blood vessels to supply tissues with oxygen and nutrients. Tumors trigger this pathway to build blood supplies, preventing necrosis and sustaining growth; and (4) hypoxia, this signature, with $200$ genes, captures the cellular survival and metabolic responses triggered by environments with low oxygen levels. As tumors outgrow their vascular supply, tumor cells activate this pathway to adapt to oxygen deprivation.
\par
The Cox models were trained on the expression of each gene set (the average expression across the genes within each molecular signature) to predict the DSS time. Supplementary Figure~\ref{fig:pathways_plot} displays the log hazard ratios/risk weights and $95$\% confidence intervals assigned to each gene set, along with their \textit{p-values}. The proliferation molecular signature is the one that contributes more to the risk score especially in TCGA-KIRC. The remaining signatures did not reach statistical significance, so no conclusions are drawn about their association with survival.
\par 
Furthermore, to take advantage of the spatial predictions at spot level, the Cox model trained with the bulk gene expression can be applied individually to each spot's predicted gene expression. This analysis should only be done in slides where the predicted pathways yielded a statistically significant \textit{p-value}, in this case TCGA-KIRC samples with the proliferation signature. This way the spatial correlation between the gene expression of each molecular signature and the risk score can be seen. Figure~\ref{fig:panel_downstream_tasks} (A) shows one of these examples, on the left the WSI along with the tumor annotation are shown, followed by the spatial expression of the proliferation signature, and on the right the spatial hazard heat map is displayed. The spatial risk scores align with the predicted spatial expression of the proliferation associated genes ($r=0.978$). These risk scores are derived from the predicted signature expression using the bulk-trained Cox model. This correlation therefore does not independently validate local risk.

\subsection{Cell-level molecular signature discovery}
\label{subsection:cell_level_molecular_signatures}

At the cellular level, model predictions were evaluated by mapping the spatial activity of MSigDB molecular signatures against subcellular ground truth in Xenium samples. Visualizing these cell-level molecular signatures might prove important in a clinical setup, for example, for the proliferation signature, the previous Sections showed that it is highly correlated with tumor related risk. Visualizing the continuous expression of this signature might help identify the cells that are ``driving'' the tumor expansion. 
\par

Figure~\ref{fig:panel_downstream_tasks} (B) shows one example of the comparison between the model's prediction of a signature's expression and the ground truth. These signature maps were compared after spatial binning and Gaussian smoothing (Section~\ref{methods:signature_maps}). The spatial PCC therefore evaluates the agreement between these signature maps, rather than the gene expression of individual cells. As with the single-cell evaluation, these comparisons are limited to the genes measured by Xenium.

\begin{figure*}[tp]
\centering
\includegraphics[width=0.82\textwidth]{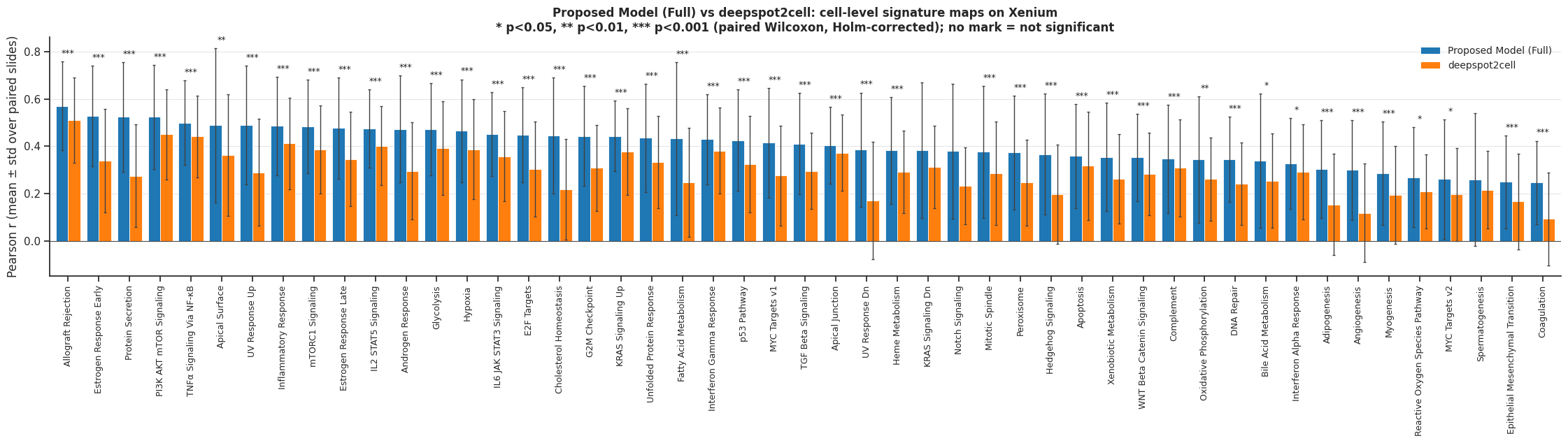}
\caption{Spatial PCC and standard deviation across paired Xenium slides of the cell-level signature maps for all hallmark gene sets, for the proposed model and DeepSpot2Cell.}
\label{fig:cell_signatures_ours_vs_deepspot2cell}
\end{figure*}

Table~\ref{tab:pcc_signatures} reports the average spatial PCC and standard deviation across slides for each signature across all Xenium slides ($N=47$) in the dataset. The model achieves a moderate spatial agreement across signatures, ranging from $r = 0.299$ for Angiogenesis to $r = 0.465$ for Hypoxia. Hypoxia and proliferation reach higher correlations ($0.465$ and $0.443$) than the gene-level cell PCC of $0.396$ (Table~\ref{tab:prediction_summary}). The two metrics are not directly comparable, since signature maps are binned and smoothed, but this is consistent with the sparsity of single-cell counts: aggregating the signal over many genes and spatial bins reduces noise and makes it easier to recover. This task shows the model's capability of predicting the gene expression at cell level and enabling molecular signatures discovery. When compared against DeepSpot2Cell under the same evaluation protocol, the proposed model achieves a higher mean correlation across all four signatures, with the largest gains in Angiogenesis ($0.299$ vs $0.118$) and Proliferation ($0.443$ vs $0.309$). Since the same slides are evaluated by both models, the differences were assessed with a paired Wilcoxon signed-rank test, which accounts for the slide-level variability. Figure~\ref{fig:cell_signatures_ours_vs_deepspot2cell} extends this comparison to all hallmark gene sets in MSigDB, the proposed model achieves a significantly higher correlation in all but three of them, for which the difference is not significant, and none of the hallmarks favors DeepSpot2Cell.

\begin{table}[tbp]
\centering
\caption{Average spatial PCC (mean $\pm$ standard deviation across slides) across the analyzed signatures in the proposed model and in DeepSpot2Cell.}
\label{tab:pcc_signatures}
\vspace{5pt}
\footnotesize
\setlength{\tabcolsep}{4pt}
\begin{tabular}{lcc}
\toprule
Signature & Proposed Model & DeepSpot2Cell \\
\midrule
Proliferation & $0.443 \pm 0.211$ & $0.309 \pm 0.181$ \\
Hypoxia       & $0.465 \pm 0.218$ & $0.387 \pm 0.211$ \\
Apoptosis     & $0.358 \pm 0.219$ & $0.318 \pm 0.229$ \\
Angiogenesis  & $0.299 \pm 0.210$ & $0.118 \pm 0.208$ \\
\bottomrule
\end{tabular}
\end{table}

\subsection{Molecular signatures localize tumor-associated regions}
\label{subsection:tcga_tumor_classification}

This task evaluates whether patch-level predictions for TCGA slides are effective for classifying tissue as tumor or non-tumor. To do this, a publicly available dataset \cite{tcga_tumor_annotations_dataset} containing tumor annotations for TCGA was used. It's important to mention that these annotations were produced by trained annotators rather than pathologists, based on the visual appearance of the tissue, and are not ``pixel-perfect'', meaning that the patch-level annotations may be coarse. Nonetheless, they provide a valuable benchmark to evaluate spot-level predictions on TCGA slides, which lack ground-truth ST.
\par
The initial analysis involves measuring the spatial correlation between predicted molecular signatures (proliferation, apoptosis, angiogenesis and hypoxia) and tumor annotations. Table~\ref{tab:pathway_metrics} reports the average AUROC and the tumor prevalence ratio across the different TCGA cohorts. These metrics were computed for each slide and averaged across all slides within each cohort. Tumor prevalence ratio measures whether patches where a molecular signature is highest are more tumor annotated compared with the cohort's baseline tumor fraction. For example, in TCGA-LUAD, the Proliferation gene set has a tumor prevalence ratio of $1.412$, indicating that patches within these hotspots are $1.412$ times more likely to be annotated as tumor than patches on that cohort overall. Moreover, in TCGA-LUAD, mainly proliferation related genes co-occur with tumor annotated patches, yielding a high AUROC ($0.711$). Additionally, for TCGA-KIRC, both hypoxia and proliferation signatures exhibit high co-occurrence with tumor annotations, mainly indicated by the high AUROC values ($0.776$ and $0.648$, respectively). Finally, in TCGA-BRCA, both proliferation and hypoxia showed the strongest AUROC values, indicating that the model successfully associates regions of tumor morphology with these gene sets.
\par
\begin{figure*}[tp]
\centering
\includegraphics[width=0.6\textwidth]{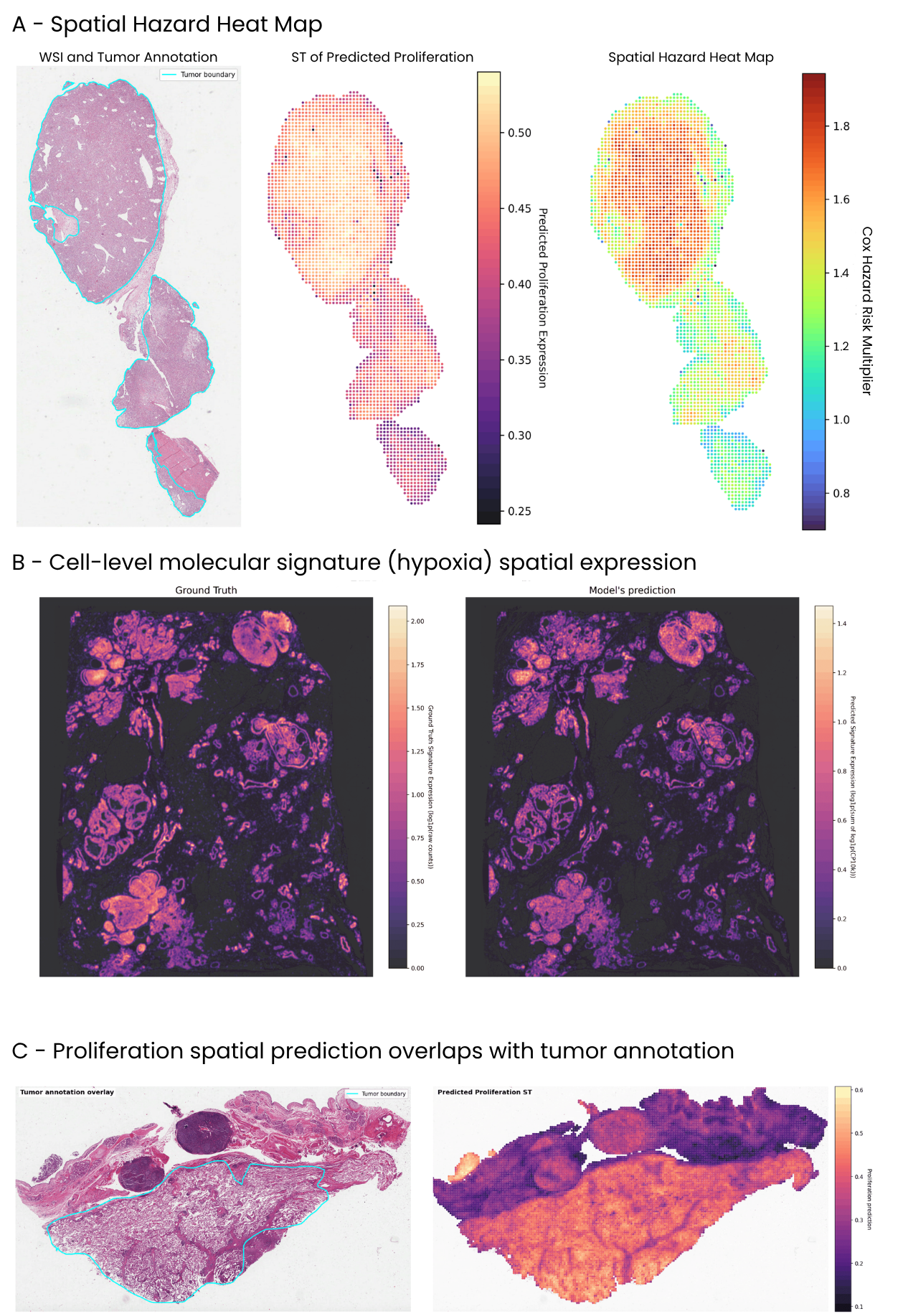}
\caption{\textbf{Downstream tasks using predicted gene expression.} \textbf{A)} WSI with tumor annotation, predicted proliferation expression, and spatial hazard heat map in a TCGA-KIRC sample. \textbf{B)} Predicted and ground-truth hypoxia signature maps in a Xenium sample, after spatial binning and Gaussian smoothing ($r=0.840$). \textbf{C)} Tumor annotation and predicted spatial proliferation expression ($\text{AUROC}=0.921$).}
\label{fig:panel_downstream_tasks}
\end{figure*}

To further illustrate these associations, Figure~\ref{fig:panel_downstream_tasks} (C) displays an example of how these molecular signatures co-occur in tumor annotated regions, supported by a high AUROC ($0.921$). Overall, these results indicate that tumor-associated gene sets (proliferation and hypoxia) consistently align with tumor annotations, showing that spot-level predictions retain tumor-related information on TCGA cohorts.

\begin{table}[tbp]
\centering
\caption{Tumor annotation and predicted molecular signature metrics by cohort. LUAD, lung adenocarcinoma; KIRC, kidney renal clear cell carcinoma; BRCA, breast invasive carcinoma.}
\label{tab:pathway_metrics}
\vspace{5pt}
\scriptsize
\setlength{\tabcolsep}{2pt}
\begin{tabular}{lccc}
\toprule
Cohort / signature & $N$ & AUROC & TPR \\
\midrule
LUAD: Prolif. & 431 & 0.711 & 1.412 \\
LUAD: Hypoxia & 431 & 0.593 & 1.131 \\
LUAD: Apopt.  & 431 & 0.525 & 0.987 \\
LUAD: Angio.  & 431 & 0.517 & 0.992 \\
\midrule
KIRC: Prolif. & 416 & 0.648 & 1.193 \\
KIRC: Hypoxia & 416 & 0.776 & 1.444 \\
KIRC: Apopt.  & 416 & 0.608 & 1.136 \\
KIRC: Angio.  & 416 & 0.579 & 1.099 \\
\midrule
BRCA: Prolif. & 850 & 0.810 & 1.910 \\
BRCA: Hypoxia & 850 & 0.728 & 1.596 \\
BRCA: Apopt.  & 850 & 0.688 & 1.485 \\
BRCA: Angio.  & 850 & 0.631 & 1.344 \\
\bottomrule
\end{tabular}

\par\vspace{2pt}
\raggedright\scriptsize
TPR: tumor prevalence ratio; Prolif.: proliferation; Angio.: angiogenesis; Apopt.: apoptosis.
\end{table}

\subsection{Tumor classification using predicted molecular signatures}

Furthermore, a three-layer Multilayer Perceptron (MLP) was trained to classify each patch as tumor or non-tumor based on predicted gene expression. The classifier was trained using three different input feature sets: (1) predicted expression for the 50 MSigDB Hallmark signatures; (2) predicted full transcriptome expression ($17{,}038$ genes); and (3) original UNI2-h patch embeddings.  Classifiers were trained and evaluated with five-fold cross-validation with splits at slide level, to avoid cross-patient contamination. Table~\ref{tab:tumor_classification} shows that both predicted transcriptomics and patch embeddings provide discriminative features for tumor classification, as reflected by high AUROC and F1-scores. Notably, the classifier performs better when using the 50 molecular signatures rather than the full transcriptome. This may happen because the individual expressions of the $17{,}038$ genes have high variance and noise, the molecular signatures act as a biologically grounded dimensionality reduction that filters out prediction noise. Using the UNI2-h embeddings outperforms using the predicted gene expression in almost all cases. This advantage should be interpreted with care: the tumor annotations were made by non-expert annotators based on the visual appearance of the tissue, which favors features that directly encode morphology, such as the PFM embeddings. Nonetheless, the $50$ signatures achieve similar results, while compressing the space into $50$ interpretable biological values.

\begin{table}[tbp]
\centering
\caption{Tumor classification performance by input type and TCGA cohort.}
\label{tab:tumor_classification}
\vspace{5pt}
\scriptsize
\setlength{\tabcolsep}{2pt}
\begin{tabular}{llccc}
\toprule
Cohort & Input & Bal. Acc. & F1-score & AUROC \\
\midrule
LUAD & 50 signatures & 0.707 & 0.707 & 0.771 \\
     & Full transcript. & 0.684 & 0.679 & 0.742 \\
     & Patch embeddings & 0.741 & 0.740 & 0.822 \\
\midrule
KIRC & 50 signatures & 0.857 & 0.861 & 0.932 \\
     & Full transcript. & 0.851 & 0.850 & 0.912 \\
     & Patch embeddings & 0.862 & 0.861 & 0.942 \\
\midrule
BRCA & 50 signatures & 0.841 & 0.838 & 0.909 \\
     & Full transcript. & 0.805 & 0.803 & 0.886 \\
     & Patch embeddings & 0.860 & 0.857 & 0.925 \\
\bottomrule
\end{tabular}

\par\vspace{2pt}
\raggedright\scriptsize
Bal. Acc.: balanced accuracy. LUAD: lung adenocarcinoma; KIRC: kidney renal clear cell carcinoma; BRCA: breast invasive carcinoma.
\end{table}

\subsection{Predictions retain tissue-specific patterns}

This task aims to assess whether the tissue-specific gene expression patterns are maintained and learned by the model. To conduct this analysis, a simple two-layer MLP was trained to classify tissue origin (lung, kidney or breast) from spot gene expression. The classifier was evaluated with five-fold cross-validation, using $105$ Visium slides ($210{,}246$ patches). The MLP was optimized on true spot expressions using cross-entropy loss and subsequently evaluated on both measured and predicted test spot expressions. Of particular interest is the difference between the MLP's predictions using the ground truth and the predicted gene expression: high similarity indicates that spot predictions retain tissue-specific patterns. To evaluate this difference, the Kullback-Leibler (KL) divergence is computed between the MLP's predicted probability vectors derived from ground-truth and the predicted expression. Specifically, for each spot, two predictions are obtained: (1) one using the true gene expression; and (2) one using the predicted gene expression by the proposed model. The MLP outputs a logit vector with three entries (one for each class: lung, kidney and breast), and a softmax function is applied to obtain class probabilities. Across all test spots, the average Kullback-Leibler (KL) divergence between the two distributions is $1.595$.

The model is further tested on $100$ randomly selected TCGA slides in each fold, using the tissue origin annotated in the TCGA metadata as ground truth.

Supplementary Table~\ref{tab:tissue_classification_transcriptomics} shows the accuracy, F1-score and AUROC averaged across the five folds; for example, the average accuracy is $0.980$ on the true Visium spot expression, $0.859$ on the predicted Visium spot expression and $0.815$ on the predicted TCGA spot expression.

In addition, the quality of the reconstructed embeddings is evaluated by assessing whether they retain tissue-specific features. The model is evaluated using five-fold cross-validation in the full dataset (TCGA, Visium and Xenium), totaling $21{,}456{,}735$ patches. The MLP is trained to classify the patch tissue origin using the original UNI2-h embeddings.

The mean KL divergence is $0.5001$. The results in Supplementary Table~\ref{tab:tissue_classification_embeddings} showing comparable accuracy, F1-score, and AUROC across both sets, confirm that the model reconstructs features from predicted ST while preserving tissue-specific characteristics.

\subsection{Ablation studies}
\label{section:ablation_sudies}

To assess the contribution of key design choices in our model, four ablated variants of the model were evaluated: (1) without feature reconstruction; (2) without consistency losses; (3) without TCGA samples; and (4) without TCGA cell level weak supervision.
Supplementary Tables~\ref{tab:visium_metrics_different_models}, \ref{tab:xenium_metrics_different_models}, and \ref{tab:tcga_metrics_different_models} report the performance metrics across benchmark models and ablated variants for the Visium, Xenium, and TCGA cohorts, respectively.

Removing the feature reconstruction objective leads to a consistent drop across most metrics on the datasets. On Visium, patch-level AUROC falls from $0.554$ to $0.548$ and the fraction of significant genes drops from $37.7\%$ to $34.6\%$. On Xenium, cell-level AUROC drops slightly ($0.649$ to $0.637$), suggesting that reconstructing the original patch embeddings encourages the model to retain morphological information that also benefits downstream expression.
\par
Removing the consistency losses degrades the structural alignment of the model's predictions. The most drastic impact is observed in Cell-to-Patch (CP) consistency on Visium, which collapses from $0.896$ to $0.222$, demonstrating that without this explicit constraint, the model fails to align both prediction levels. Patch-to-Slide (PS) consistency also drops across all datasets, including a substantial decrease on TCGA ($0.893$ to $0.691$). Consequently, this lack of structural regularization harms overall bulk predictive performance on the spatially resolved datasets, dropping bulk PCC from $0.774$ to $0.726$ on Visium and from $0.797$ to $0.779$ on Xenium.
\par
Removing TCGA from training has a more mixed effect: bulk PCC improves substantially on both Visium ($0.774$ to $0.875$) and Xenium ($0.797$ to $0.831$), likely because the model is trained exclusively, at slide-level, with pseudo-bulk labels from these cohorts. However, this comes at the cost of reduced feature reconstruction quality (Visium: $0.634$ to $0.522$; Xenium: $0.566$ to $0.453$), indicating that TCGA's scale and diversity still contribute to feature reconstruction.
\par
Finally, ablating the cell-level weak supervision from TCGA reveals its important role in fine-grained spatial localization. In Xenium, patch-level AUROC drops ($0.703$ to $0.674$), while cell-level AUROC also drops slightly ($0.649$ to $0.640$). This demonstrates that the cell-level information extracted from TCGA cohort, although it offers only weak supervision to the model, seems beneficial for guiding the model to accurately resolve expression patterns at the patch and cell-levels, even when evaluating on entirely different spatial platforms.

\section{Discussion}
In this work, a Multi-Scale GNN was developed to predict gene expression from WSIs at three levels: bulk, spatial and single-cell. The model combines data from TCGA, Visium and Xenium samples, using direct supervision when gene expression measurements are available and weak supervision through consistency losses. This allows the model to predict $17{,}038$ genes across lung, kidney and breast tissues, while learning from datasets with different levels of data granularity.

The results show that the model performs best at slide level, reaching an average PCC of $0.881$ on TCGA. At patch level, the average PCC was $0.509$ on Visium and $0.786$ on Xenium. Predicting cells was more challenging, with an average PCC of $0.396$ on Xenium. Compared with DeepSpot2Cell, our model obtained higher cell-level PCC and AUROC, while DeepSpot2Cell was better at patch level, in both PCC and AUROC. These results support the use of a shared model for the three prediction tasks, although performance remains uneven across genes and tissues.

The downstream tasks showed that the predictions retain tissue-specific and tumor-related information. Predicted molecular signatures supported tumor localization and classification, while the cell-derived signature maps partially reproduced spatial patterns observed in Xenium. Using 50 molecular signatures for tumor classification performed better than using the full predicted gene expression, showing the value of grouping genes into interpretable biological processes. Predicted bulk expression also supported survival analysis in kidney and breast cancer, although performance was lower than with measured expression in all three cohorts.

The ablation studies showed that the consistency losses are important for keeping predictions aligned across scales. Feature reconstruction and weak supervision from TCGA cells also helped several prediction metrics. However, adding TCGA did not improve every task: removing these samples increased the bulk PCC on Visium and Xenium. This shows that combining datasets provides useful information, but also introduces cohort specific differences that the model must accommodate.

These findings have some limitations. Single-cell predictions could only be directly evaluated for the genes measured by Xenium, and TCGA spatial predictions could not be compared with ground truth. Batch effects, small cohorts, cell segmentation errors and memory constraints also affected the study. Furthermore, differences in the supervision available to each model limit the conclusions drawn from the benchmark comparisons. Nevertheless, this work provides a framework for combining bulk, spatial and single-cell data to obtain molecular predictions from histopathology images.

Future work should evaluate the model on larger and more balanced datasets, including samples from independent institutions. This would help determine how well the model handles changes in tissue preparation, imaging and sequencing platforms. Broader single-cell measurements would also allow more of the predicted genes to be validated.

Another area of improvement is addressing batch effects directly during training. While algorithms like Harmony can align representations in low-dimensional spaces, they cannot be easily applied to the full transcriptome during end-to-end learning. Future iterations could incorporate methods that explicitly penalize the model for learning hospital or scanner-specific features, encouraging the model to focus strictly on biological patterns rather than technical differences between cohorts.

On the architectural level, the current pipeline only passes information bottom-up, from cells to patches and then to the slide. Incorporating top-down information flow would allow global tissue context to guide local patch and cell predictions, which could help resolve ambiguous local features. Additionally, the cell feature extraction pipeline could be refined. Instead of using a frozen patch-level foundation model on upsampled cell crops, fine-tuning the visual backbone or using encoders pre-trained specifically on single-cell morphology could provide more informative input embeddings for the cell graph.

\section{Methods}

\subsection{Dataset and experimental design}
\label{methods:dataset}

The multi-modal dataset created comprises several individual datasets, spanning across three different organs: lung, kidney and breast. The dataset aggregates bulk RNA-seq data from TCGA studies, and ST and scT data from several different studies all aggregated by the HEST-1k dataset \cite{hest}.  For this study, from HEST-1k, $152$ cases were selected, lung had a total of $60$ cases ($36$ Visium and $24$ Xenium), kidney had $64$ cases ($62$ Visium and $2$ Xenium), while breast had $28$ cases ($7$ Visium and $21$ Xenium). The ST data comes from the Visium samples, on the other hand, scT data is present in Xenium samples that were acquired using the Xenium In Situ Platform which is commercialized by 10x Genomics.

When it comes to the bulk gene expression data, there is much more data available. Publicly available TCGA datasets for lung adenocarcinoma (TCGA-LUAD) \cite{tcga_luad_2014}, clear cell renal cell carcinoma (TCGA-KIRC) \cite{tcga_kirc_2013} and breast invasive carcinoma (TCGA-BRCA) \cite{tcga_brca}, were analyzed and obtained via the NCI Genomic Data Commons portal. In the TCGA-LUAD project we used $465$ cases, this is, $465$ pairs of DX (diagnostic) slides and bulk gene expression, in TCGA-KIRC $510$ cases were used and in TCGA-BRCA there were $1052$ cases used. In TCGA projects the number of sequenced genes is $60{,}660$. Out of the $2027$ TCGA samples, $268$ of them were randomly selected to have their cells detected, segmented and classified (with Classpose) and then their features extracted with UNI2-h.  For these slides, during training, the model performs weak supervision at cell level through the consistency losses. At inference time, the model is able to also predict at cell-level for these slides, effectively generating novel virtual scT data.

In terms of training strategy, the model was trained using five-fold cross-validation with splits at slide level, to avoid cross-patient contamination (there is only one WSI for each patient), meaning that the test WSIs/patients were never seen during training. Moreover, due to memory constraints of the hardware used, a hard limit of $300{,}000$ cells per WSI at test time was used. For slides exceeding this limit, $300{,}000$ cells were randomly sampled for inference, and the remaining cells were not processed. The evaluated datasets contained $1{,}724$ TCGA, $105$ Visium and $47$ Xenium slides.

\subsection{WSI segmentation and feature extraction}
\label{methods:images}
For the HEST-1k slides with ST, each patch is centered around a spot; consequently, the patch's gene expression is approximated by the gene expression in the spot. Throughout this work, the terms spot and patch are used interchangeably.

Tissue segmentation involves deciding which parts of the image are tissue or background and removing the background. Trident \cite{zhang2025standardizing} was the segmentation tool used.  Trident also allows automatic patching of the segmented tissue: it extracts the coordinates of each patch based on two specified parameters, patch size and magnification (here, $224 \times 224\,\text{px}$ at $20\times$ magnification). With the coordinates of each patch in the WSI, the patch images can be extracted and fed into the PFM (UNI2-h in this work) for feature extraction.

Following WSI segmentation and patching, the next step in the pipeline is the segmentation and classification of individual cells. The model used for this task was Classpose \cite{classpose}, which is designed to perform simultaneous instance segmentation and classification of cells in histopathology images. A Classpose model was trained on the PanNuke dataset \cite{pannuke, pannuke_ext} to predict five cell classes: neoplastic, inflammatory, connective, dead and epithelial. The data was randomly split into $80\%$ for training and $20\%$ for testing, and training used the default Classpose parameters with uncertainty weighting, a batch size of $4$ and $200$ epochs. To compare with CellViT \cite{cellvit}, the currently available model trained on PanNuke, separate models were also trained on the three original PanNuke training folds. The average over the three validation folds shows that Classpose outperforms CellViT in panoptic quality (PQ) for all cell types and matches or exceeds it in detection quality (DQ) (Supplementary Table~\ref{tab:classpose_vs_cellvit}). Test performance on the $20\%$ hold-out set closely matches the 3-fold average: the PQ/DQ for neoplastic, inflammatory, connective, dead and epithelial cells is $0.601/0.711$, $0.546/0.652$, $0.444/0.557$, $0.319/0.423$ and $0.595/0.738$, respectively.

For each detected cell, the $64 \times 64 \text{px}$ (at the maximum magnification, typically $20 \times$) crops around each detected cell are extracted. These $64\text{px} \times 64 \text{px}$ crops are then upsampled to $224\text{px} \times 224 \text{px}$, and their features are extracted with UNI2-h, producing cell embeddings.

A particularly relevant PFM that was extensively used is \textit{UNI2-h} \cite{chen2024uni}, a Vision Transformer (ViT) designed specifically for large-scale histopathology representation learning. This model takes as input a $224 \times 224$px patch image and outputs a $1536 \times 1$ embedding.

\subsection{Transcriptomic measurements and normalization}
\label{methods:expression}

Single-Cell Transcriptomics (scT) reveals how individual cells express genes. The data used in the study was acquired using the Xenium In Situ Platform, which is commercialized by 10x Genomics. Contrary to the Visium, also commercialized by 10x Genomics, the Xenium is an imaging-based technology which provides subcellular spatial resolution. This technology enables the detection of mRNA transcripts with their exact 2D position within the tissue, which, when paired with the cell segmentation from Classpose, allows a gene expression to be assigned to each cell. This is done by counting the mRNA transcripts that intersect each cell (the area of the cells is defined by the segmentation mask inferred by Classpose), essentially assigning which mRNA transcripts belong to each cell.
\par

Despite its high reliability, the main drawback of using Xenium samples is their limited gene panel, the samples used from the HEST-1k dataset only contain $300$ to $600$ genes, depending on the sample, which is fairly limited when compared to the $10{,}000$ to $33{,}000$ genes sequenced by Visium. Furthermore, it is worth mentioning that scT can be converted into ST data for compatibility. The HEST-1k dataset already handles this by creating ``pseudo-Visium'' spots derived from the sub-cellular mRNA transcripts detected in each $55 \times 55\,\mu\mathrm{m}$ spot.

Raw count data can vary from gene to gene, from zero up to millions of counts. Such high variance is problematic for deep learning models to learn from. These problems introduce the need for a normalization strategy that is invariant to the library size (total number of mRNA transcripts detected) and that compresses the variance. The strategy used was to apply the $\log(1+\text{CP10k})$ normalization, which compresses the variance to the range of $0-8$ and makes gene expression comparable between different samples.

\subsection{Construction of the graphs}
\label{methods:graphs}

The principle used for constructing the graphs is that neighboring patches/cells can share important information and by giving each node the spatial context from its neighboring nodes, the prediction made for that node will be better when compared to not giving spatial information about the neighboring nodes. To this end, an algorithm was designed to build graphs based on the neighborhood and distance between the nodes. The algorithm starts by computing the k-Nearest Neighbors (k-NN) for each node.

Patch-level graphs were built using a dynamic-radius neighborhood with $k=6$ (radius factor $1.5$) for HEST slides and $k=8$ for TCGA slides, while cell-level graphs used a pruned $k=6$ neighborhood with a distance multiplier of $1.5$.

\subsection{Multi-Scale GNN architecture}
\label{methods:architecture}
Supplementary Figure~\ref{fig:multiscalegnnarchitecture} depicts the architecture of the proposed model. The architecture consists of four main components that work together: (1) the cell prediction branch, which includes the \textit{Cell GAT Backbone}, \textit{Cell Head}, \textit{Cell Attention} and \textit{Fusion Gate}; (2) the patch prediction branch, which includes the \textit{Patch GAT Backbone}, \textit{Patch Head} and \textit{Patch Attention} ; (3) the slide prediction branch, that is comprised by the \textit{Slide Head}; and (4) the reconstruction branch which is only composed by the \textit{Reconstruction Head}.

In this model's architecture the information always flows from the smaller scale to the higher scale, information from the cell branch flows to the patch branch and information from the patch branch flows to the slide branch.

The cell branch is responsible for receiving the cells' feature graph and predicting each cell's gene expression. The cell feature graph contains one node for each detected cell, whose initial embedding was extracted using UNI2-h. The \textit{Cell GAT backbone} is responsible for performing the attention mechanism between graph nodes and essentially modeling the cell-cell interactions. This produces a refined feature graph where each cell is represented by an updated embedding of size $1 \times 1024$ enriched by the neighborhood nodes through the attention mechanism. Each cell's refined embedding is provided to the \textit{Cell Head}, which regresses the gene expression vector.
\par
In order to enrich the patch branch with cell-level information, the \textit{Cell Attention} and \textit{Fusion Gate} are employed. The \textit{Cell Attention} layer computes attention scores across the refined cell embeddings, which are normalized using the \textit{Per Patch Softmax} layer to weight the importance of each cell to its parent patch.  Consequently, each patch will be assigned an embedding that represents the information gathered by the cell branch. The \textit{Fusion Gate} is used to map this embedding from the cell feature space into the patch feature space, where it is added to the patch's refined embedding in the patch branch.
\par
During training, only 25 patches per WSI are randomly selected to retain the underlying cell graph.    Since most patches in the WSI will not have the cell's graph, the embedding that results from the \textit{Per Patch Softmax} layer will be all zeros, to preserve the all zeros embedding the first layer of the fusion gate has $\text{bias}=0$. This results in the cell's information embedding being all zeros and not affecting the patch's refined embedding.

 The \textit{Patch GAT Backbone} performs two \textit{GATConv} operations, meaning that the message passing mechanism performs a two neighbor hop attention.  A skip connection is added to prevent oversmoothing and preserve the original features of each node. The refined patch embeddings are enriched with the cell's information embedding, after which the \textit{Patch Head} regresses a gene expression vector for each node.
\par
The \textit{Patch Attention} layer is responsible for weighing the importance of each patch to create an embedding that represents the slide embedding.
The slide branch is composed just by the \textit{Slide Head}, which takes as input the slide embedding output by the \textit{Patch Attention} layer and regresses the bulk gene expression vector.

The reconstruction branch is composed by the \textit{Reconstruction Head}, this layer is responsible for reconstructing the original patch features from the predicted patch gene expressions. This is basically a parallel objective to generate patch-level feature representations from ST data.

\subsection{Multi-Scale aggregation}
\label{methods:aggregation}
Mathematically, normalized ($\log(1+\text{CP10k})$) transcriptomic distributions cannot be linearly aggregated in the normalized space, as most state-of-the-art models suggest \cite{mosby, nonchev2025deepspot2cell}. Summing or averaging normalized transcriptomics profiles would severely distort the normalization properties. In order to maintain biological consistency across all prediction levels, in the training stage, the proposed hierarchical aggregation algorithm implements the following pipeline: (1) It reverts the predicted multi-level gene expressions out of their logarithmic domains back into the library-size corrected ($\text{CP10k}$) space; (2) It aggregates these library-size corrected expressions across different levels using library-size weighted pooling, effectively going back into the raw count space and summing the gene expressions; and (3) It maps the resulting raw count profiles back into the normalized space.

In cases where the true cells' library sizes are unavailable (TCGA and Visium samples), the model approximates the aggregation by simply averaging the $\text{CP10k}$ values of the cell within each patch. This approximation assumes that all cells have the same library size, which is biologically incorrect, but necessary when cell's library sizes are unobserved.

For training samples that do not have patch-level ground-truth (TCGA) and thus the patch library sizes are unknown, the model performs an average of the patch-level $\text{CP10k}$ values.

\subsection{Loss function and optimization}
\label{methods:loss}
In order to train the Multi-Scale GNN across bulk (TCGA), spot-level (Visium) and single-cell (Xenium) datasets, a hybrid weakly and fully supervised co-training strategy was developed. To ensure balanced learning across all modalities, the training loop dynamically mixes data from all three datasets. 

Because training is conducted on a multi-modal dataset, available ground-truth labels differ across cohorts, resulting in sample-dependent loss formulations. For TCGA samples only the slide-level ground truth is available, so only the \textit{Slide Head} gets direct supervision, on the other hand, the \textit{Patch Head} and \textit{Cell Head} receive weak supervision through the consistency losses. For Visium samples, the ground truth is available at spot-level, and the slide-level ground truth is inferred by the spot-level labels (by summing the raw counts of all genes in all spots), creating the pseudo-bulk gene expression. For these training samples there is direct supervision at slide and patch-level, while also having all the consistency losses. Finally, for Xenium samples, the ground truth is available at sub-cellular level, the cell ground truth is computed by counting the transcripts inside each detected cell, as discussed in Section~\ref{methods:expression}, there are also pseudo-spots and pseudo-bulk gene expression ground truths available. For these samples, all the regression heads get direct supervision, along with the weak supervision from the consistency terms.

A hybrid Pearson-MSE loss was used, mixing squared-error minimization with MSE and structural pattern alignment with PCC:
\begin{equation}
\label{eq:pearson_mse}
\begin{aligned}
\mathcal{L}_{\text{Pearson-MSE}}(P,T) = \delta \cdot \operatorname{MSE}(P,T) + (1-\delta)\cdot \bigl(1-\operatorname{PCC}(P,T)\bigr).
\end{aligned}
\end{equation}
In this formulation, $P$ and $T$ represent the predicted and ground-truth gene expression profiles, and $\delta=0.3$ balances the two components.

For the direct supervision tasks, the loss components were dynamically balanced using an uncertainty weighting scheme proposed in \cite{uncertainty_weights}. The total loss function integrates the direct supervision targets along with the cross-level consistency constraints:
\begin{equation}
\label{eq:loss_function}
\begin{gathered}
    \mathcal{L}_{\text{total}} = \frac{1}{\sigma_{\text{slide}}^2} \mathcal{L}_{\text{slide}} + \frac{1}{\sigma_{\text{patch}}^2} \mathcal{L}_{\text{patch}} + \frac{1}{\sigma_{\text{cell}}^2} \mathcal{L}_{\text{cell}} \\
    {} + \lambda_{\text{CP}}(t) \mathcal{L}_{\text{CP}} + \lambda_{\text{PS}}(t) \mathcal{L}_{\text{PS}} + \omega(t) \mathcal{L}_{\text{recon}} + \log(\sigma_{\text{slide}}\sigma_{\text{patch}}\sigma_{\text{cell}})
\end{gathered}
\end{equation}
Here, $\sigma_{\text{slide}}$, $\sigma_{\text{patch}}$ and $\sigma_{\text{cell}}$ are learnable parameters. The learned weights should not be interpreted as the task's uncertainty/variance, but as coarse task-level balancing parameters that improve optimization by balancing supervised tasks with different noise levels and different learning dynamics.

The direct supervision terms use the Pearson-MSE loss. $\mathcal{L}_{\text{CP}}$ ensures consistency between pooled single-cell predictions and their corresponding patches, while $\mathcal{L}_{\text{PS}}$ aligns patch-level predictions with global slide predictions; both also use the Pearson-MSE loss. $\mathcal{L}_{\text{recon}}$ drives the reconstruction of the original patch features using a pure MSE loss. The regularization weights scale linearly from zero up to their target values of $0.01$ for both $\lambda_{\text{CP}}$ and $\lambda_{\text{PS}}$, and $0.1$ for $\omega$, over the first $50$ epochs.

Optimization used AdamW (learning rate $1\times10^{-3}$, weight decay $1\times10^{-4}$) with a \texttt{ReduceLROnPlateau} scheduler (factor $0.5$, patience of $5$ epochs, minimum learning rate $1\times10^{-6}$), mixed-precision training with gradient scaling, and gradient-norm clipping at $1.0$. Training proceeded for a maximum of $50$ epochs, along with a batch size of $3$ WSIs and gradient accumulation of $16$ steps, with an effective batch size of $48$ WSIs.

\subsection{Baseline models and benchmarks}
\label{methods:baselines}
In order to establish a baseline, two other models (DeepSpot2Cell and STFlow) were trained and tested on our dataset. Due to the architecture of these models, which only have supervision at the patch-level, they can only be trained with WSIs which have ground-truth at spot-level (Visium and Xenium). For this reason DeepSpot2Cell and STFlow were only trained on our Visium and Xenium data, TCGA samples were not considered. Nonetheless, for direct comparison, these models were trained on the exact same splits as our model, excluding the TCGA samples. Both baselines used the same UNI2-h embeddings, common gene panel, and random seed ($42$) as our model.

\subsection{Evaluation metrics}
\label{methods:metrics}

The Pearson Correlation Coefficient (PCC) is used throughout this work as the primary metric for assessing how well the predicted expression profiles align with the ground truth, both across genes within a sample and across samples for a given gene. Since gene expression prediction can also be framed as a discrimination problem, that is, distinguishing spatial regions or cells of high versus low expression, the Area Under the Receiver Operating Characteristic (AUROC) is used to complement the PCC by evaluating this property directly.

This metric was designed for classification tasks, but as suggested in \cite{stnet} it can be applied to continuous gene expressions as we have in ST data. To leverage AUROC for evaluating the model's ability to capture distinct spatial profiles, the continuous ground-truth expression values are binarized into ``high'' or ``low'' expression.
\par
Specifically, in the evaluation pipeline, for each gene across all spots in each WSI, the $25^{\text{th}}$ percentile ($p_{25}$) and the $75^{\text{th}}$ percentile ($p_{75}$) of the gene's true spatial expression distribution are computed. For each gene, the spots are then partitioned into high expression ($H$) and low expression ($L$) subsets. $H$ contains spots where the spot's gene expression is greater or equal to $p_{75}$, while $L$ contains spots where the gene expression is lower or equal to $p_{25}$. Intermediate spots between $p_{25}$ and $p_{75}$ are masked and excluded from the computation. 
\par
The AUROC is then calculated using these discrete subsets $H$ and $L$ against the model's continuous predicted expression values for the unmasked spots. Finally, the individual gene AUROC scores are averaged to yield a per WSI spatial discrimination metric. A high AUROC in this context means that the model consistently predicts higher expression values in regions that genuinely correspond to the top quartile of gene expression.

To assess the statistical reliability of these correlations across thousands of genes simultaneously, the False Discovery Rate (FDR) is used to control for false discoveries. In practice, the FDR is implemented using the Benjamini-Hochberg (BH) procedure \cite{fdr}. First, the PCC between the true and predicted spatial expressions is calculated for each gene on every test slide. A one-sided binomial test is then applied to the signs of these slide-level correlations, producing one $p$-value per gene. Genes for which the correlation cannot be computed are excluded (happens in cases where the measured or predicted expression is constant). Genes satisfying $p^{\mathrm{BH}} < 0.05$ are classified as having a significantly predicted spatial expression. To allow direct comparison, FDR values are reported as percentages of valid genes per fold.

\subsection{Cell-level molecular signature maps}
\label{methods:signature_maps}

To compare predicted and ground-truth signature distributions, each WSI was partitioned into a grid of $2{,}000 \times 2{,}000$ spatial bins. Within each bin, the gene expression of the signature was computed by summing the predicted expression of each cell within the bin. On the other hand, for the ground truth, the signature expression associated with each bin is computed by summing the number of raw transcripts associated to the signature inside the bin. Moreover, since single-cell data is extremely sparse and noisy, spatial smoothing was applied using a Gaussian filter. The comparison between the proposed model and DeepSpot2Cell was assessed with a paired Wilcoxon signed-rank test across slides.

\section{Conclusion}
This work presented a Multi-Scale GNN that predicts bulk, spatial and single-cell transcriptomics directly from WSIs. By combining TCGA, Visium and Xenium samples in a hybrid training strategy, with direct supervision where gene expression measurements are available and weak supervision through consistency losses, a single model learns from datasets with different levels of granularity while keeping its predictions aligned across scales.

The model performs best at the slide level and moderately at the patch level, while single-cell prediction remains the most challenging task; at this level, it achieved higher correlation and AUROC than DeepSpot2Cell. Performance remains uneven across genes and tissues, and single-cell predictions could only be directly evaluated for the genes measured by Xenium. Nevertheless, the downstream tasks showed that the predicted profiles retain tissue-specific and tumor-related information, supporting survival analysis, tumor localization and classification, and cell-level molecular signature maps.

Although spatial and single-cell transcriptomics do not yet have an established clinical application, this work helps advance computational tools that make them accessible from routine histopathology images, which may prove valuable as their practical medical utility develops.

\FloatBarrier
\begingroup
\small
\setlength{\emergencystretch}{3em}

\endgroup

\clearpage
\onecolumn
\section*{Supplementary material}
\setcounter{figure}{0}
\setcounter{table}{0}
\renewcommand{\thefigure}{S\arabic{figure}}
\renewcommand{\thetable}{S\arabic{table}}
\renewcommand{\theHfigure}{S\arabic{figure}}
\renewcommand{\theHtable}{S\arabic{table}}

\subsection*{TCGA spatial predictions and tumor classification}
\begin{figure}[htbp]
\centering
\includegraphics[width=1.0\textwidth]{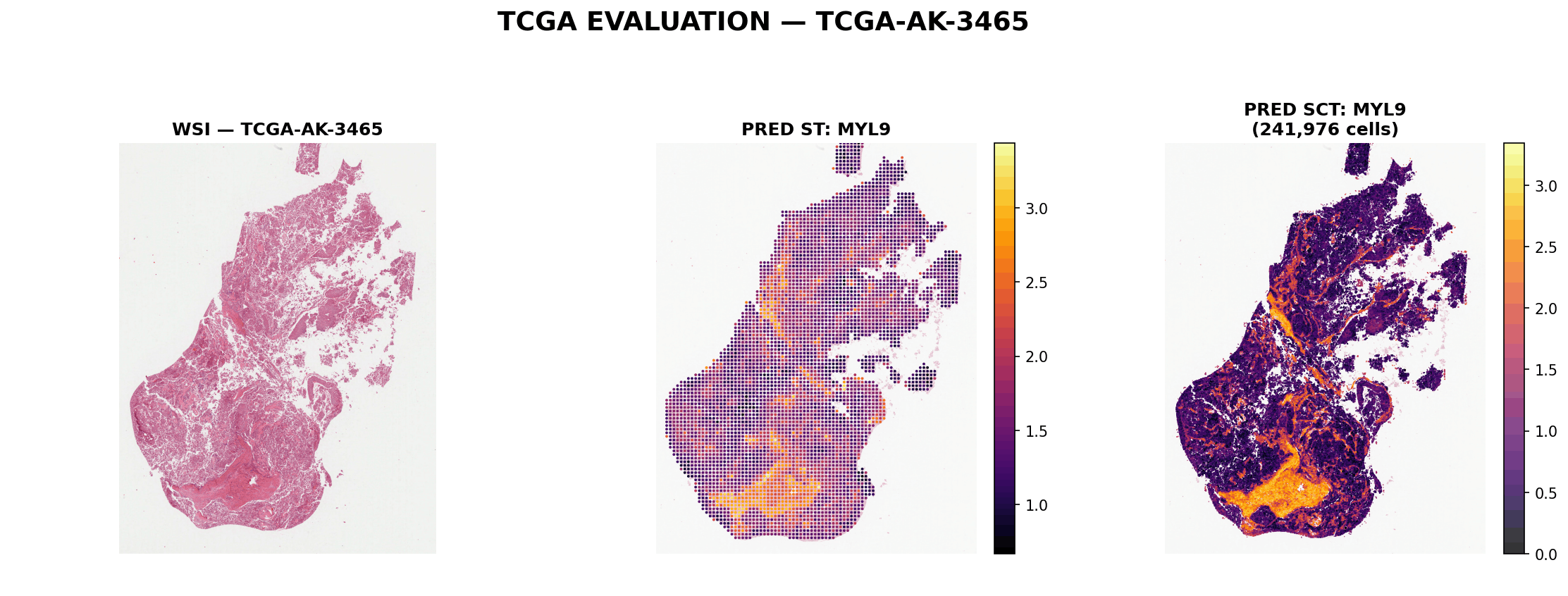}
\caption{ST and scT predictions for TCGA slide AK-3465, gene \textit{MYL9}. Left to right: WSI, predicted spot expression, predicted cell expression.}
\label{fig:tcga_ak_3465}
\end{figure}

In Figure~\ref{fig:tcga_ak_3465}, the model produces spatially coherent expression patterns that align with tissue morphology visible in the corresponding H\&E image, with the finer-grained cell-level predictions recovering the same regional expression trends identified at the spot level.

\begin{figure}[p]
\centering
\includegraphics[width=1.0\textwidth]{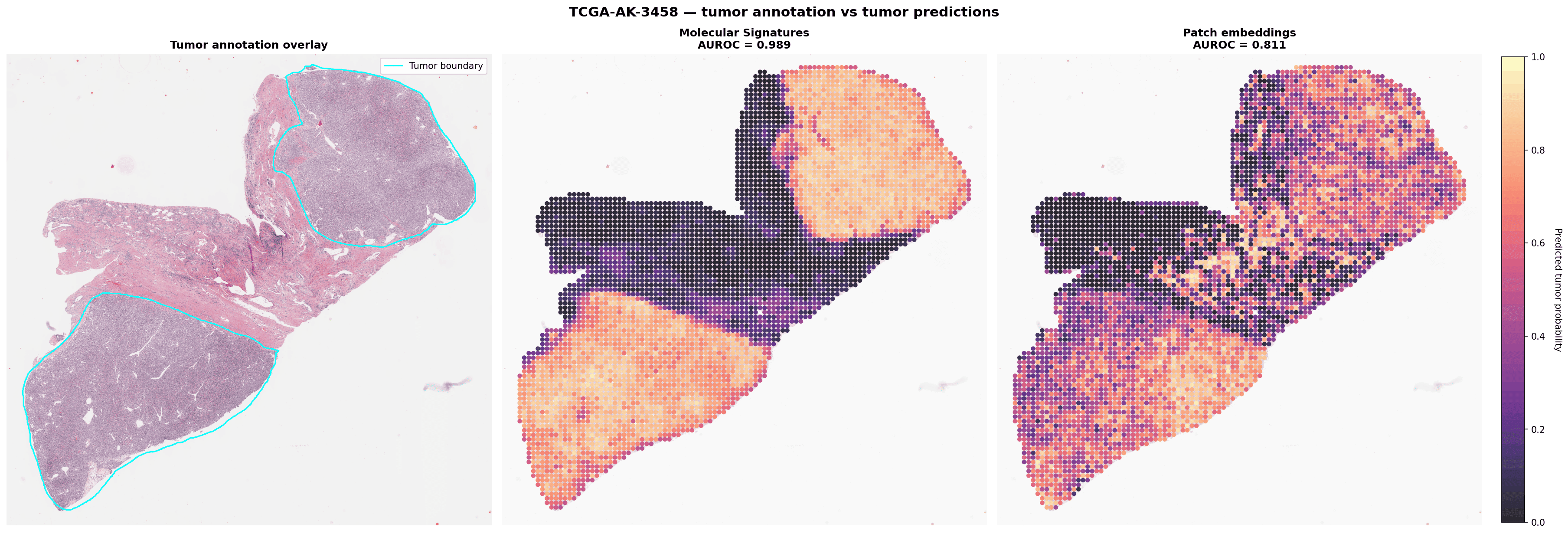}
\caption{Comparison between the tumor annotations and the predicted tumor probability by each feature set in TCGA slide AK-3458.}
\label{fig:tumor_annotation_vs_preds_2}
\end{figure}

In the single-slide example in Figure~\ref{fig:tumor_annotation_vs_preds_2}, predictions based on the 50 molecular signatures achieve higher discrimination ($\text{AUROC} = 0.989$) than the original patch embeddings ($\text{AUROC} = 0.811$).

\subsection*{Visual features correlate with gene expression}
The clustering analysis is described in Section~\ref{section:visual_features}. The $100$ principal components of the patch features explain $69.82\%$ of the variance, while the $300$ principal components of the gene expression vectors explain $25.76\%$ of the variance.

The cluster purity measures, among patches in the same patch-feature cluster, the percentage that map to the same gene expression cluster. ARI measures pair-wise consistency, adjusted for agreements that would happen purely by chance. A score of $0$ indicates random clustering, while $1$ indicates identical clustering. Finally, NMI measures information overlap between the two cluster assignments, ranging from $0$ (no relationship) to $1$ (perfect predictability).

\subsection*{Cell type encodes gene expression}
\begin{figure}[htbp]
\centering
\includegraphics[width=1.0\textwidth]{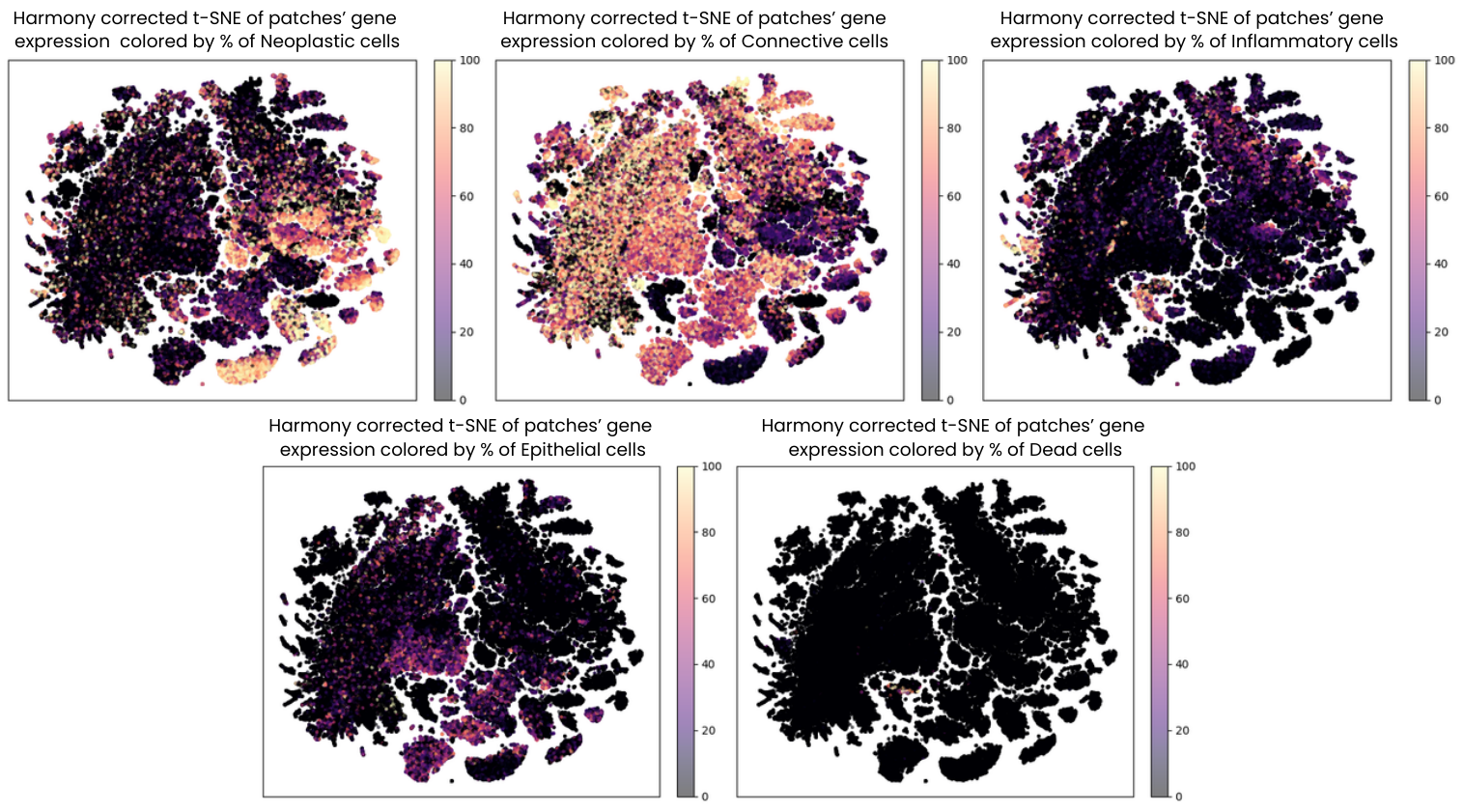}
\caption{Harmony corrected t-SNEs of patches' gene expression profiles colored by percentage of each cell type (neoplastic, connective, inflammatory, epithelial, dead) (across $250{,}000$ randomly selected patches).}
\label{fig:st_colored_w_cell_percentage}
\end{figure}

In the t-SNE of the Harmony-corrected patches' gene expression of Visium slides, each point is colored by the percentage of cell types present in the corresponding patch (Figure~\ref{fig:st_colored_w_cell_percentage}). Connective cells, the most common cell type, appear frequently in most clusters.

\begin{figure}[htbp]
\centering
\includegraphics[width=1.0\textwidth]{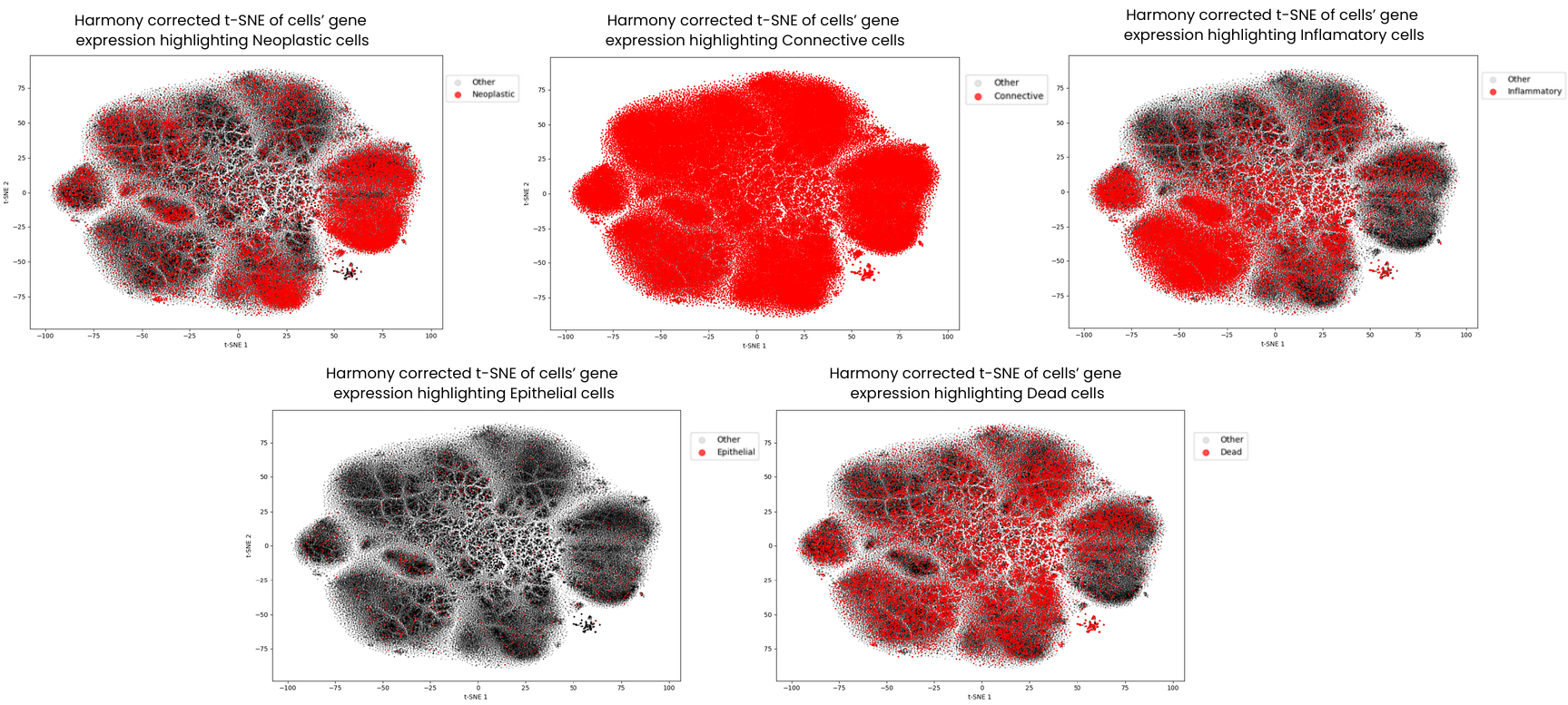}
\caption{Harmony corrected t-SNEs of cells' gene expression profiles highlighting cell type (neoplastic, connective, inflammatory, epithelial, dead) (across $250{,}000$ randomly selected cells).}
\label{fig:sct_colored_by_cell_type}
\end{figure}

The same analysis can be done at the cell level on the Xenium slides available in the HEST-1k dataset. To do this, a gene expression is assigned to each cell, as described in Section~\ref{methods:expression}. Out of the available cells, $250{,}000$ were selected randomly for the analysis. In Figure~\ref{fig:sct_colored_by_cell_type}, it is possible to see the t-SNEs of the Harmony-corrected cells' gene expression, highlighting each cell type. There is not a clear cluster for each cell type: for connective cells, for example, it appears that there is not a specific gene expression profile. On the other hand, for neoplastic and inflammatory cells, there is clearly a certain gene expression profile associated with them.

\subsection*{Multi-Scale GNN architecture}
\begin{figure}[p]
\centering
\includegraphics[width=0.88\textwidth]{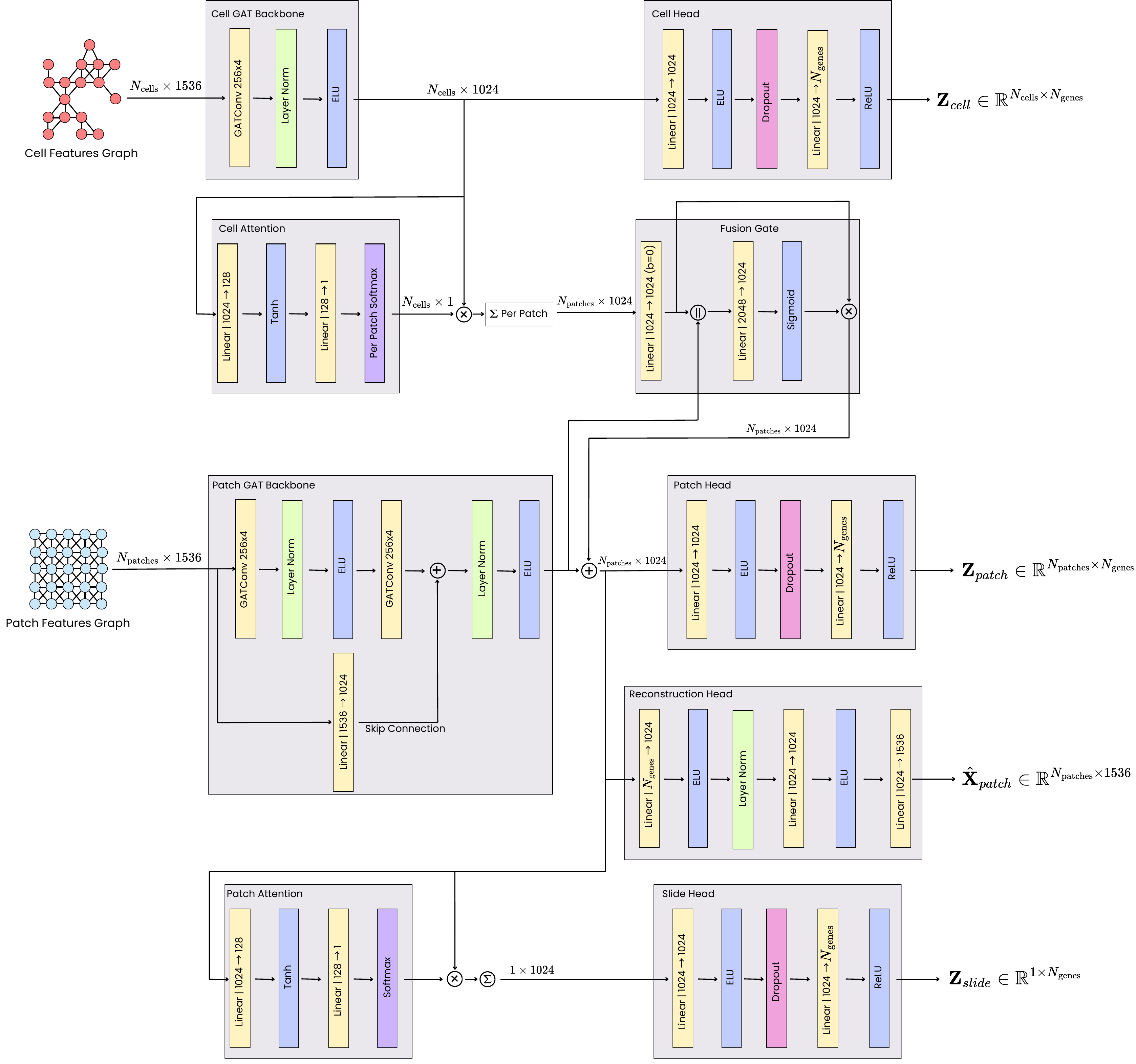}
\caption{Architecture of the Multi-Scale Graph Neural Network (GNN).}
\label{fig:multiscalegnnarchitecture}
\end{figure}

Figure~\ref{fig:multiscalegnnarchitecture} depicts the architecture described in Section~\ref{methods:architecture}. Each WSI can have up to millions of detected cells; even with a small batch size of WSIs, it would be unfeasible to hold every cell's embedding in GPU memory. Considering that each of the $25$ patches selected during training might have $5$--$50$ cells, each training sample will have $125$--$1{,}250$ cells. This makes each WSI have a considerably smaller memory footprint and enables much faster training.

\subsection*{Multi-Scale aggregation}
The aggregation strategy and the approximations used when library sizes are unavailable are described in Section~\ref{methods:aggregation}. The following equations detail the cell-to-patch and patch-to-slide aggregation.

\paragraph{Cell-level to patch-level aggregation.}
Let $\mathbf{Z}_{\text{cell}} \in \mathbb{R}^{N_{cells} \times N_{genes}}$ represent the predicted cell gene expression matrix generated by the cell-level head, where $N_{cells}$ is the total number of detected cells and $N_{genes}$ is the number of target genes. Every individual cell index $i \in \{1, \dots, N_{cells}\}$ belongs to a parent patch index $j \in \{1, \dots, N_{patches}\}$.

In order to aggregate the cell predictions into their corresponding patch, the model first reverts the normalized prediction ($\mathbf{z}_{\text{cell}, i}$) to raw counts ($\text{Raw}_{\text{cell}, i}$). To do this, the model first reverts the logarithm to obtain the $\text{CP10k}$ value, and then un-scales it using the cell's true library size ($L_{\text{cell},i}$):
\begin{equation}
\text{Raw}_{\text{cell}, i} = \left( \exp(\mathbf{z}_{\text{cell}, i}) - 1 \right) \times \frac{L_{\text{cell}, i}}{10^4}
\end{equation}

Subsequently, the raw counts of all cells mapped to patch $j$ are summed to compute the total raw gene expression of the patch ($\text{Raw}_{\text{patch},j}$). The aggregated library size for the patch is the sum of its underlying cells' library sizes ($L_{\text{patch}, j} = \sum L_{\text{cell}, i}$).

Finally, the original normalization is re-applied to the raw expression to yield the predicted cell-aggregated patch prediction:
\begin{equation}
\mathbf{z}_{\text{patch\_from\_cells}, j} = \log\left(1 + \frac{\text{Raw}_{\text{patch}, j}}{L_{\text{patch}, j}} \times 10^4\right)
\end{equation}

\paragraph{Patch-level to slide-level aggregation.}
The predicted patch level gene expressions ($\mathbf{z}_{\text{patch}, j}$) are converted to raw counts using their respective spot library sizes ($L_{\text{patch},j}$):
\begin{equation}
\text{Raw}_{\text{patch}, j} = \left( \exp(\mathbf{z}_{\text{patch}, j}) - 1 \right) \times \frac{L_{\text{patch}, j}}{10^4}
\end{equation}

The raw counts from all patches are summed to approximate the bulk gene expression at slide-level ($\text{Raw}_{\text{slide}}$). The original normalization is re-applied using the slide library size to produce the final patch derived bulk gene expression:
\begin{equation}
\mathbf{z}_{\text{slide\_from\_patches}} = \log\left(1 + \frac{\text{Raw}_{\text{slide}}}{L_{\text{slide}}} \times 10^4\right)
\end{equation}

\subsection*{Loss weighting}
The loss function and optimization settings are described in Section~\ref{methods:loss}. The $\log(\sigma_{\text{slide}}\sigma_{\text{patch}}\sigma_{\text{cell}})$ term in Equation~\ref{eq:loss_function} is a regularization term to prevent the model from driving the weights to zero in order to minimize the loss.

The reason why the consistency and reconstruction terms were not weighted using the learned weights was because, as experimentally observed, the model becomes over-confident on these tasks, assigning them higher weights and neglecting the direct supervision losses. This happens because these tasks are ``easier'' and the model just maintains the consistency between the predictions instead of trying to learn the directly supervised tasks. For this reason, these weights were kept small and deterministic.

\subsection*{Baseline models and benchmarks}
The benchmark design is described in Section~\ref{methods:baselines}. Both baseline models were trained using their respective published architectures and default hyperparameters. DeepSpot2Cell used an ensemble of $8$ Phi/Rho sub-networks (hidden size $512$, dropout $0.3$), each aggregating information from a spot's $k=8$ nearest spatial neighbors. It was trained with AdamW (learning rate $1\times10^{-4}$, weight decay $1\times10^{-6}$) for up to $50$ epochs, batch size $256$ spots, with early stopping after $5$ epochs without improvement. STFlow was configured as a $4$-layer spatial transformer denoiser (hidden dimension $128$, edge dimension $128$, $4$ attention heads, dropout $0.2$) trained under a conditional-flow-matching objective with a zero-inflated negative-binomial prior, using AdamW (learning rate $5\times10^{-4}$, weight decay $1\times10^{-6}$) for up to $100$ epochs with early stopping (patience of $20$ epochs), gradient-norm clipping at $1.0$, a batch size of $1$ slide.

\subsection*{Ablation studies}
The four ablated variants are described in Section~\ref{section:ablation_sudies}. Removing feature reconstruction tests whether this self-supervised signal improves the quality of the learned representations used for gene expression prediction. Removing the patch to slide and cell to patch consistency terms tests whether this constraint is helpful for the model. Removing TCGA slides from training tests whether the additional scale and tissue diversity provided by this large, spatially-unresolved dataset benefits performance on the spatially-resolved Visium and Xenium benchmarks. Removing TCGA cell level weak supervision tests whether the extra information at cell-level from TCGA is beneficial to the model.
\FloatBarrier

\begin{table}[H]
\centering
\caption{Proposed model's metrics for the Visium Dataset}
\label{tab:visium_metrics}
\resizebox{\textwidth}{!}{
\begin{tabular}{lcccccccc}
\toprule
\multirow{2}{*}{\textbf{Cohort}} & \multirow{2}{*}{\textbf{Total $N$}} & \textbf{Bulk} & \textbf{Patch} & \textbf{Feat. Recon.} & \textbf{PS Cons.} & \textbf{CP Cons.} & \textbf{Patch} & \textbf{Signif. Genes (Patch)} \\
 & & \textbf{PCC} & \textbf{PCC} & \textbf{Cos. Sim.} & \textbf{PCC} & \textbf{PCC} & \textbf{AUROC} & \textbf{(FDR < 0.05)} \\
\midrule
Overall & 105 & $0.774 \pm 0.146$ & $0.509 \pm 0.026$ & $0.634 \pm 0.009$ & $0.777 \pm 0.142$ & $0.896 \pm 0.016$ & $0.554 \pm 0.004$ & $37.7\% \pm 5.9\%$ \\
Breast  & 7   & $0.752 \pm 0.080$ & $0.583 \pm 0.053$ & $0.481 \pm 0.068$ & $0.816 \pm 0.082$ & $0.892 \pm 0.028$ & $0.567 \pm 0.014$ & $-$ \\
Kidney  & 62  & $0.742 \pm 0.216$ & $0.544 \pm 0.024$ & $0.672 \pm 0.040$ & $0.753 \pm 0.216$ & $0.907 \pm 0.014$ & $0.536 \pm 0.005$ & $14.0\% \pm 9.7\%$ \\
Lung    & 36  & $0.844 \pm 0.071$ & $0.430 \pm 0.035$ & $0.609 \pm 0.020$ & $0.819 \pm 0.042$ & $0.874 \pm 0.031$ & $0.586 \pm 0.018$ & $13.8\% \pm 13.8\%$ \\
\bottomrule
\end{tabular}
}
\end{table}

\begin{table}[H]
\centering
\caption{Proposed model's metrics for the Xenium Dataset}
\label{tab:xenium_metrics}
\resizebox{\textwidth}{!}{
\begin{tabular}{lccccccccccc}
\toprule
\multirow{2}{*}{\textbf{Cohort / Subtype}} & \multirow{2}{*}{\textbf{Total $N$}} & \textbf{Bulk} & \textbf{Patch} & \textbf{Cell} & \textbf{Feat. Recon.} & \textbf{PS Cons.} & \textbf{CP Cons.} & \textbf{Patch} & \textbf{Signif. Genes (Patch)} & \textbf{Cell} \\
 & & \textbf{PCC} & \textbf{PCC} & \textbf{PCC} & \textbf{Cos. Sim.} & \textbf{PCC} & \textbf{PCC} & \textbf{AUROC} & \textbf{(FDR < 0.05)} & \textbf{AUROC} \\
\midrule
Overall & 47 & $0.797 \pm 0.057$ & $0.786 \pm 0.041$ & $0.396 \pm 0.021$ & $0.566 \pm 0.044$ & $0.858 \pm 0.044$ & $0.779 \pm 0.029$ & $0.703 \pm 0.011$ & $8.0\% \pm 14.0\%$ & $0.649 \pm 0.011$ \\
\quad Connective & - & - & - & $0.367 \pm 0.019$ & - & - & - & - & - & $0.616 \pm 0.003$ \\
\quad Dead & - & - & - & $0.330 \pm 0.019$ & - & - & - & - & - & $0.613 \pm 0.018$ \\
\quad Epithelial & - & - & - & $0.406 \pm 0.023$ & - & - & - & - & - & $0.626 \pm 0.022$ \\
\quad Inflammatory & - & - & - & $0.340 \pm 0.017$ & - & - & - & - & - & $0.596 \pm 0.012$ \\
\quad Neoplastic & - & - & - & $0.456 \pm 0.036$ & - & - & - & - & - & $0.614 \pm 0.016$ \\
\addlinespace
Breast & 21 & $0.785 \pm 0.040$ & $0.775 \pm 0.019$ & $0.399 \pm 0.026$ & $0.503 \pm 0.024$ & $0.868 \pm 0.018$ & $0.781 \pm 0.010$ & $0.736 \pm 0.010$ & $-$ & $0.678 \pm 0.027$ \\
\quad Connective & - & - & - & $0.332 \pm 0.031$ & - & - & - & - & - & $0.622 \pm 0.009$ \\
\quad Dead & - & - & - & $0.275 \pm 0.063$ & - & - & - & - & - & $0.643 \pm 0.027$ \\
\quad Epithelial & - & - & - & $0.402 \pm 0.032$ & - & - & - & - & - & $0.626 \pm 0.029$ \\
\quad Inflammatory & - & - & - & $0.308 \pm 0.042$ & - & - & - & - & - & $0.607 \pm 0.033$ \\
\quad Neoplastic & - & - & - & $0.475 \pm 0.023$ & - & - & - & - & - & $0.592 \pm 0.018$ \\
\addlinespace
Kidney & 2  & $0.572 \pm 0.000$ & $0.517 \pm 0.000$ & $0.169 \pm 0.000$ & $0.504 \pm 0.000$ & $0.778 \pm 0.000$ & $0.602 \pm 0.000$ & $0.567 \pm 0.000$ & $-$ & $0.578 \pm 0.000$ \\
\quad Connective & - & - & - & $0.161 \pm 0.000$ & - & - & - & - & - & $0.572 \pm 0.000$ \\
\quad Epithelial & - & - & - & $0.164 \pm 0.000$ & - & - & - & - & - & $0.440 \pm 0.000$ \\
\quad Inflammatory & - & - & - & $0.168 \pm 0.000$ & - & - & - & - & - & $0.537 \pm 0.000$ \\
\quad Neoplastic & - & - & - & $0.156 \pm 0.000$ & - & - & - & - & - & $0.513 \pm 0.000$ \\
\addlinespace
Lung & 24 & $0.828 \pm 0.057$ & $0.823 \pm 0.042$ & $0.416 \pm 0.024$ & $0.622 \pm 0.057$ & $0.851 \pm 0.062$ & $0.796 \pm 0.028$ & $0.685 \pm 0.015$ & $15.3\% \pm 34.1\%$ & $0.629 \pm 0.020$ \\
\quad Connective & - & - & - & $0.417 \pm 0.027$ & - & - & - & - & - & $0.617 \pm 0.006$ \\
\quad Dead & - & - & - & $0.368 \pm 0.030$ & - & - & - & - & - & $0.600 \pm 0.018$ \\
\quad Epithelial & - & - & - & $0.433 \pm 0.023$ & - & - & - & - & - & $0.623 \pm 0.028$ \\
\quad Inflammatory & - & - & - & $0.380 \pm 0.027$ & - & - & - & - & - & $0.595 \pm 0.014$ \\
\quad Neoplastic & - & - & - & $0.464 \pm 0.045$ & - & - & - & - & - & $0.639 \pm 0.013$ \\
\bottomrule
\end{tabular}
}
\end{table}

\begin{table}[H]
\centering
\caption{Proposed model's metrics for the TCGA Dataset}
\label{tab:tcga_metrics}
\resizebox{\textwidth}{!}{
\begin{tabular}{lcccccc}
\toprule
\multirow{2}{*}{\textbf{Cohort}} & \multirow{2}{*}{\textbf{Total $N$}} & \textbf{Bulk} & \textbf{Feat. Recon.} & \textbf{PS Cons.} & \textbf{Bulk} & \textbf{Signif. Genes (Bulk)} \\
 & & \textbf{PCC} & \textbf{Cos. Sim.} & \textbf{PCC} & \textbf{AUROC} & \textbf{(FDR < 0.05)} \\
\midrule
Overall & $1{,}724$ & $0.881 \pm 0.010$ & $0.637 \pm 0.014$ & $0.893 \pm 0.015$ & $0.661 \pm 0.003$ & $72.6\% \pm 16.0\%$ \\
Breast  & $866$  & $0.875 \pm 0.012$ & $0.615 \pm 0.015$ & $0.891 \pm 0.018$ & $0.608 \pm 0.007$ & $53.1\% \pm 12.3\%$ \\
Kidney  & $423$  & $0.886 \pm 0.025$ & $0.682 \pm 0.014$ & $0.881 \pm 0.030$ & $0.586 \pm 0.015$ & $31.0\% \pm 20.0\%$ \\
Lung    & $435$  & $0.888 \pm 0.004$ & $0.637 \pm 0.016$ & $0.907 \pm 0.008$ & $0.602 \pm 0.005$ & $42.2\% \pm 4.8\%$ \\
\bottomrule
\end{tabular}
}
\end{table}

\begin{figure}[htbp]
\centering
\includegraphics[width=\textwidth]{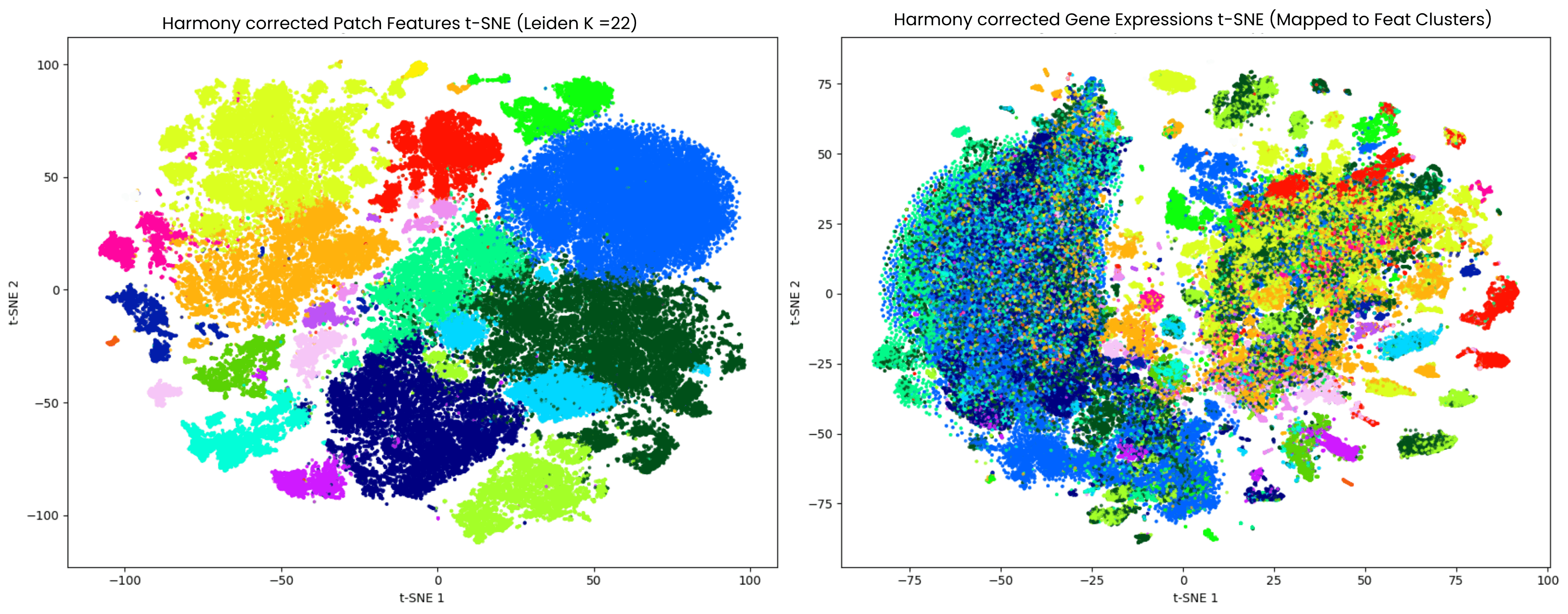}
\caption{Left: Harmony corrected t-SNE of clustered patch features. Right: Harmony corrected t-SNE of gene expression profiles, colored according to the cluster assignment of the corresponding patch features (across $150{,}000$ randomly selected patches).}
\label{fig:tsne_genes_features_clustered_harmonized}
\end{figure}

\begin{figure}[htbp]
\centering
\includegraphics[width=1.0\textwidth]{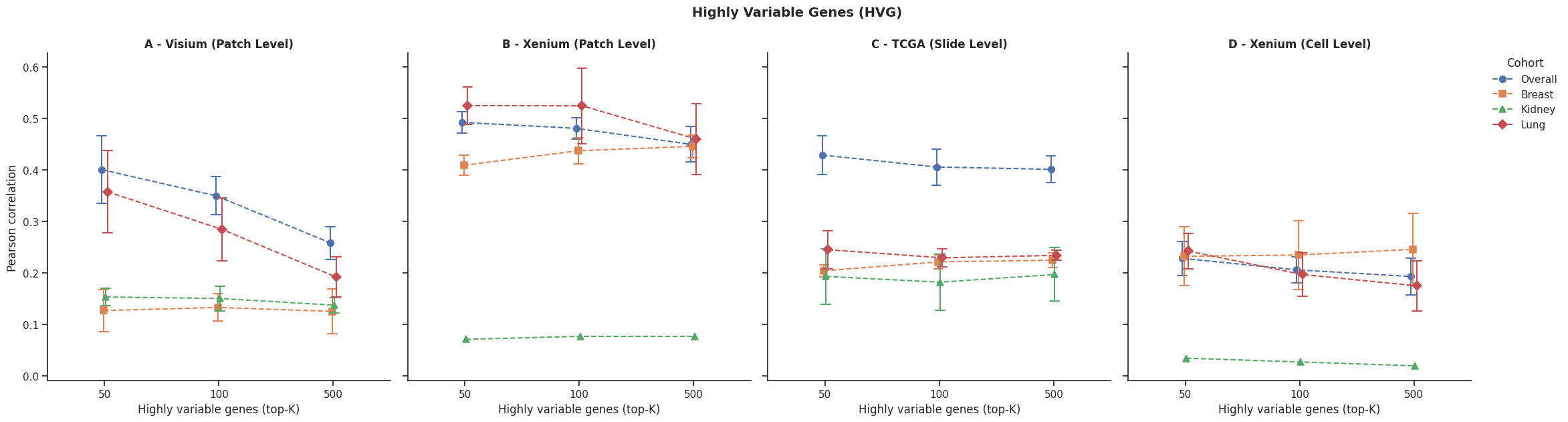}
\caption{Gene-wise mean PCC in each dataset against the number of top Highly Variable Genes (HVGs) for the proposed model.}
\label{fig:gene_wise_pcc_per_dataset_hvg}
\end{figure}

\begin{figure}[htbp]
\centering
\includegraphics[width=1.0\textwidth]{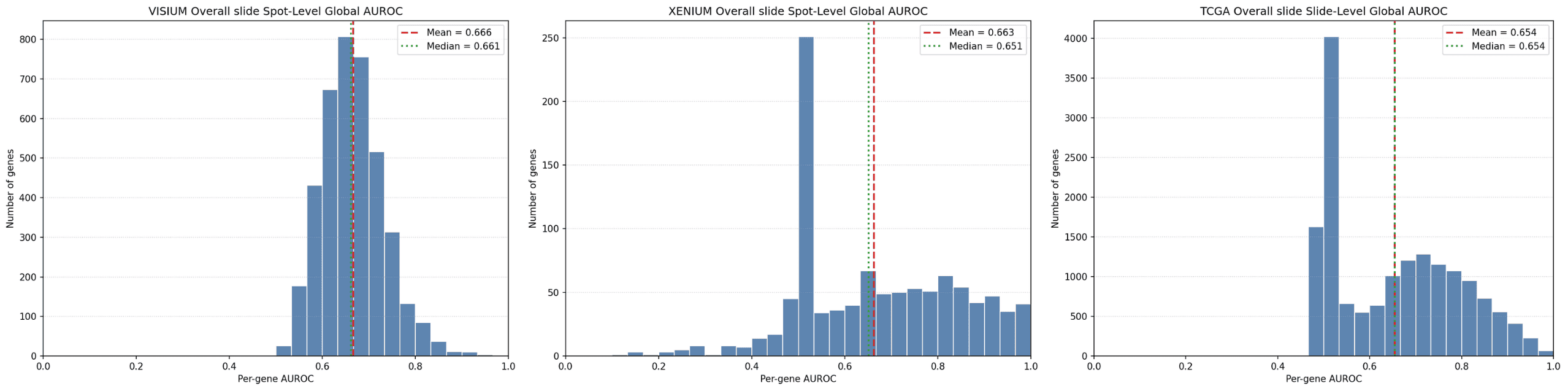}
\caption{Histograms of gene-wise AUROC in each dataset for the proposed model.}
\label{fig:aurocs_histogram}
\end{figure}

\begin{figure}[htbp]
  \centering
  \subfloat[Log hazard ratios for each molecular signature to predict DSS time in TCGA-LUAD.\label{fig:pathways_luad}]{%
    \includegraphics[width=0.9\textwidth]{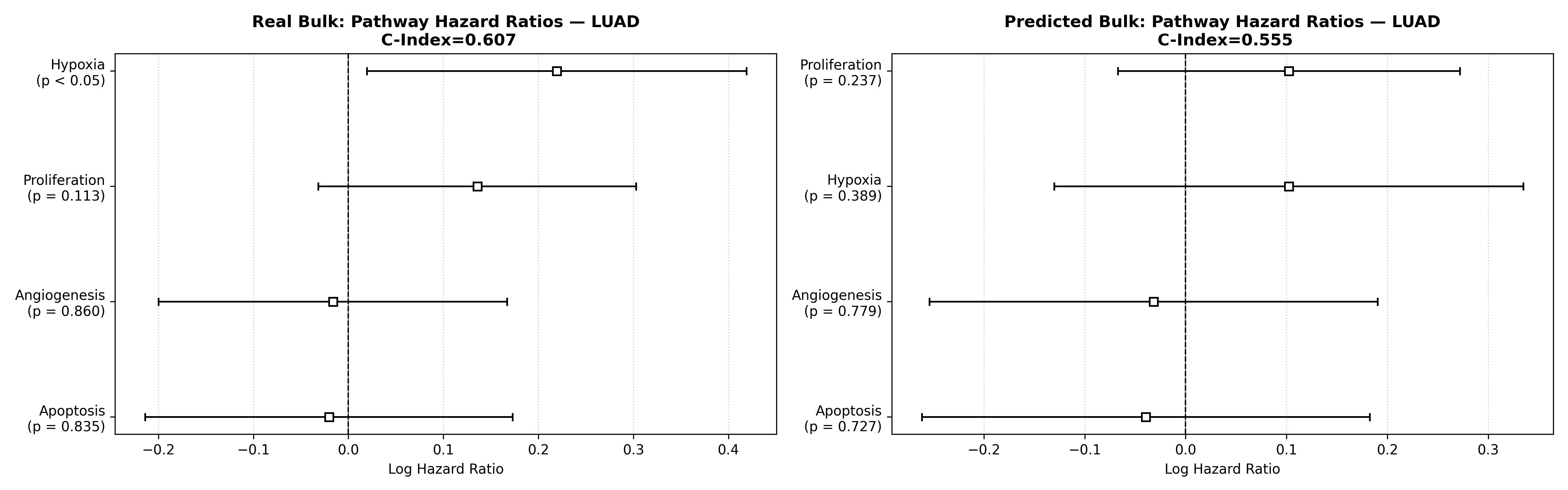}%
  }\hfill
  \subfloat[Log hazard ratios for each molecular signature to predict DSS time in TCGA-KIRC.\label{fig:pathways_kirc}]{%
    \includegraphics[width=0.9\textwidth]{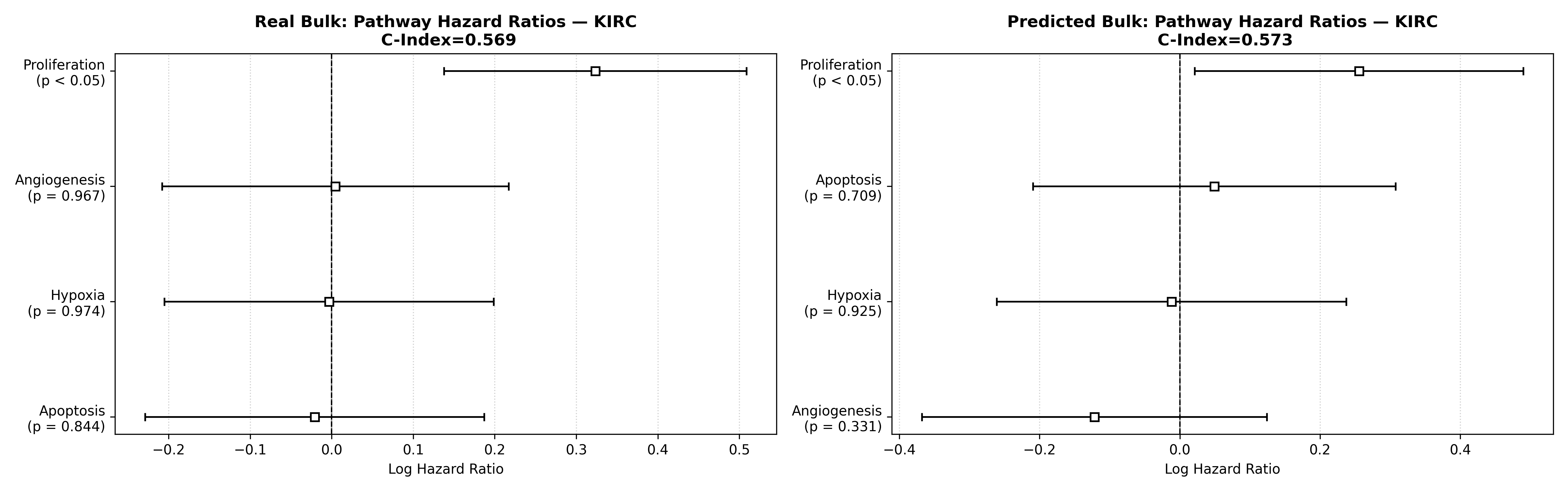}%
  }\hfill
  \subfloat[Log hazard ratios for each molecular signature to predict DSS time in TCGA-BRCA.\label{fig:pathways_brca}]{%
    \includegraphics[width=0.9\textwidth]{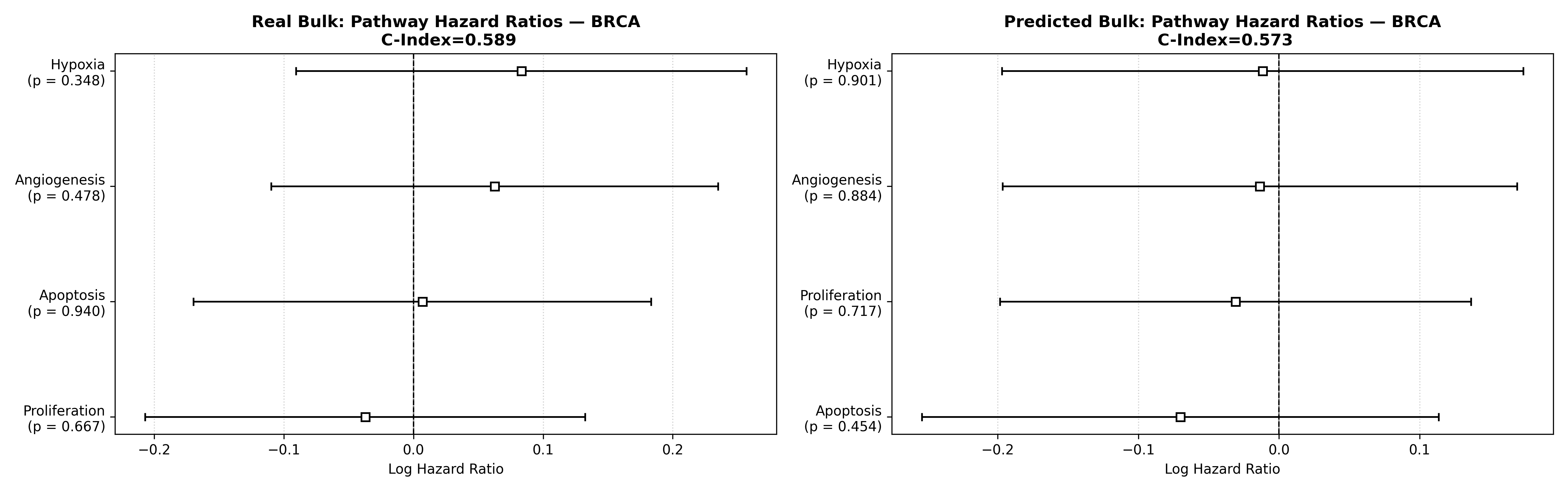}%
  }
  \caption{Log hazard ratios for each molecular signature to predict DSS time in TCGA cohorts.}
  \label{fig:pathways_plot}
\end{figure}

\begin{table}[htbp]
\centering
\caption{MLP's classification metrics across the true and predicted spot's gene expression.}
\label{tab:tissue_classification_transcriptomics}
\resizebox{\textwidth}{!}{%
\begin{tabular}{lccc}
\toprule
\textbf{Test Set} & \textbf{\begin{tabular}[c]{@{}c@{}} Average Accuracy\\\end{tabular}} & \textbf{\begin{tabular}[c]{@{}c@{}} Average F1-score\\\end{tabular}} & \textbf{\begin{tabular}[c]{@{}c@{}} Average AUROC\\\end{tabular}} \\ 
\midrule
Visium true spot's gene expression       & $0.980 \pm 0.025$ & $0.975 \pm 0.028$ & $0.994 \pm 0.009$ \\
Visium predicted spot's gene expression  & $0.859 \pm 0.068$ & $0.804 \pm 0.101$ & $0.942 \pm 0.056$ \\
TCGA predicted spot's gene expression  & $0.815 \pm 0.029$ & $0.774 \pm 0.038$ & $0.910 \pm 0.022$ \\
\bottomrule
\end{tabular}
}
\end{table}

\begin{table}[htbp]
\centering
\caption{MLP's classification metrics across the original and reconstructed spot's embedding.}
\label{tab:tissue_classification_embeddings}
\begin{tabular}{lccc}
\toprule
\textbf{Test Set} & \textbf{\begin{tabular}[c]{@{}c@{}} Accuracy\\\end{tabular}} & \textbf{\begin{tabular}[c]{@{}c@{}}F1-score\\\end{tabular}} & \textbf{\begin{tabular}[c]{@{}c@{}}AUROC\\\end{tabular}} \\ 
\midrule
Original spot's embedding       & $0.924 \pm 0.007$ & $0.915 \pm 0.008$ & $0.984 \pm 0.005$ \\
Reconstructed spot's embedding  & $0.902 \pm 0.009$ & $0.893 \pm 0.011$ & $0.981 \pm 0.005$ \\
\bottomrule
\end{tabular}
\end{table}

\begin{figure}[htbp]
\centering
\includegraphics[width=1.0\textwidth]{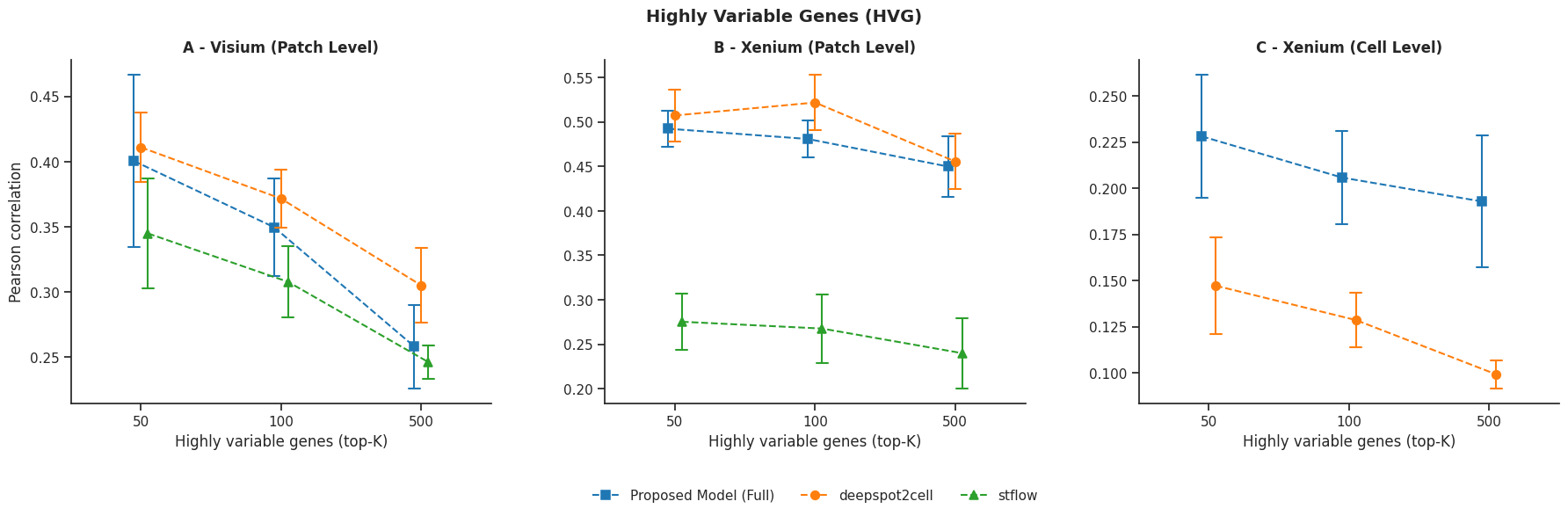}
\caption{Gene-wise mean PCC in each model against the number of top HVGs.}
\label{fig:sota_models_comparison_hvg}
\end{figure}

\begin{table}[H]
\centering
\caption{Metrics for the Visium Dataset (Overall) across different models.}
\label{tab:visium_metrics_different_models}
\resizebox{\textwidth}{!}{
\begin{tabular}{lcccccccc}
\toprule
\multirow{2}{*}{\textbf{Model}} & \multirow{2}{*}{\textbf{Total $N$}} & \textbf{Bulk} & \textbf{Patch} & \textbf{Feat. Recon.} & \textbf{PS Cons.} & \textbf{CP Cons.} & \textbf{Patch} & \textbf{Signif. Genes (Patch)} \\
 & & \textbf{PCC} & \textbf{PCC} & \textbf{Cos. Sim.} & \textbf{PCC} & \textbf{PCC} & \textbf{AUROC} & \textbf{(FDR < 0.05)} \\
\midrule
Proposed Model  (Full)            & $105$ & $0.774 \pm 0.146$ & $0.509 \pm 0.026$ & $0.634 \pm 0.009$ & $0.777 \pm 0.142$ & $0.896 \pm 0.016$ & $0.554 \pm 0.004$ & $37.7\% \pm 5.9\%$ \\
Proposed Model (w/o Consistency)  & $105$ & $0.726 \pm 0.178$ & $0.512 \pm 0.027$ & $0.606 \pm 0.036$ & $0.740 \pm 0.183$ & $0.222 \pm 0.007$ & $0.555 \pm 0.004$ & $43.5\% \pm 9.3\%$ \\
Proposed Model (w/o Feat. Recon.) & $105$ & $0.744 \pm 0.159$ & $0.501 \pm 0.030$ & $-$               & $0.757 \pm 0.168$ & $0.881 \pm 0.029$ & $0.548 \pm 0.002$ & $34.6\% \pm 12.5\%$ \\
Proposed Model (w/o TCGA)          & $105$ & $0.875 \pm 0.017$ & $0.510 \pm 0.025$ & $0.522 \pm 0.033$ & $0.924 \pm 0.019$ & $0.867 \pm 0.030$ & $0.549 \pm 0.011$ & $36.4\% \pm 16.0\%$ \\
Proposed Model (w/o TCGA cells)    & $105$ & $0.794 \pm 0.085$ & $0.506 \pm 0.028$ & $0.559 \pm 0.063$ & $0.814 \pm 0.090$ & $0.850 \pm 0.027$ & $0.548 \pm 0.005$ & $37.3\% \pm 4.7\%$ \\
DeepSpot2Cell           & $105$ &  -                & $0.535 \pm 0.028$ & $-$               & -                 & $-$               & $0.590 \pm 0.004$ & $55.1\% \pm 4.4\%$ \\
STFlow                  & $105$ & -                 & $0.519 \pm 0.032$ & $-$               & -                 & $-$               & $0.575 \pm 0.009$ & $48.7\% \pm 6.6\%$ \\
\bottomrule
\end{tabular}
}
\end{table}

\begin{table}[H]
\centering
\caption{Metrics for the Xenium Dataset (Overall) across different models.}
\label{tab:xenium_metrics_different_models}
\resizebox{\textwidth}{!}{
\begin{tabular}{lcccccccccc}
\toprule
\multirow{2}{*}{\textbf{Model}} & \multirow{2}{*}{\textbf{Total $N$}} & \textbf{Bulk} & \textbf{Patch} & \textbf{Cell} & \textbf{Feat. Recon.} & \textbf{PS Cons.} & \textbf{CP Cons.} & \textbf{Patch} & \textbf{Signif. Genes (Patch)} & \textbf{Cell} \\
 & & \textbf{PCC} & \textbf{PCC} & \textbf{PCC} & \textbf{Cos. Sim.} & \textbf{PCC} & \textbf{PCC} & \textbf{AUROC} & \textbf{} & \textbf{AUROC} \\
\midrule
Proposed Model   (Full)           & $47$ & $0.797 \pm 0.057$ & $0.786 \pm 0.041$ & $0.396 \pm 0.021$ & $0.566 \pm 0.044$ & $0.858 \pm 0.044$ & $0.779 \pm 0.029$ & $0.703 \pm 0.011$ & $8.0\% \pm 14.0\%$  & $0.649 \pm 0.011$ \\
Proposed Model (w/o Consistency)  & $47$ & $0.779 \pm 0.022$ & $0.770 \pm 0.054$ & $0.390 \pm 0.029$ & $0.564 \pm 0.078$ & $0.837 \pm 0.032$ & $0.734 \pm 0.032$ & $0.691 \pm 0.027$ & $8.7\% \pm 15.5\%$  & $0.638 \pm 0.014$ \\
Proposed Model (w/o Feat. Recon.) & $47$ & $0.810 \pm 0.045$ & $0.777 \pm 0.049$ & $0.384 \pm 0.032$ & $-$ & $0.863 \pm 0.044$ & $0.741 \pm 0.056$ & $0.686 \pm 0.015$ & $8.0\% \pm 15.1\%$ & $0.637 \pm 0.013$ \\
Proposed Model (w/o TCGA)          & $47$ & $0.831 \pm 0.036$ & $0.771 \pm 0.045$ & $0.398 \pm 0.028$ & $0.453 \pm 0.056$ & $0.924 \pm 0.009$ & $0.780 \pm 0.042$ & $0.669 \pm 0.031$ & $7.8\% \pm 15.1\%$  & $0.648 \pm 0.015$ \\
Proposed Model (w/o TCGA cells)    & $47$ & $0.764 \pm 0.079$ & $0.759 \pm 0.041$ & $0.391 \pm 0.020$ & $0.519 \pm 0.096$ & $0.817 \pm 0.115$ & $0.739 \pm 0.035$ & $0.674 \pm 0.022$ & $7.3\% \pm 13.0\%$  & $0.640 \pm 0.009$ \\
DeepSpot2Cell           & $47$ & -                 & $0.760 \pm 0.055$ & $0.349 \pm 0.034$ & $-$               & $0.984 \pm 0.011$ & $0.726 \pm 0.028$ & $0.764 \pm 0.010$ & $8.8\% \pm 16.1\%$ & $0.523 \pm 0.013$ \\
STFlow                  & $47$ & -                 & $0.679 \pm 0.088$ & $-$               & $-$               & $1.000 \pm 0.000$ & $-$               & $0.634 \pm 0.019$ & $4.5\% \pm 10.1\%$  & $-$ \\
\bottomrule
\end{tabular}
}
\end{table}

\begin{table}[H]
\centering
\caption{Metrics for the TCGA Dataset (Overall) across different models.}
\label{tab:tcga_metrics_different_models}
\resizebox{\textwidth}{!}{
\begin{tabular}{lcccccc}
\toprule
\multirow{2}{*}{\textbf{Model}} & \multirow{2}{*}{\textbf{Total $N$}} & \textbf{Bulk} & \textbf{Feat. Recon.} & \textbf{PS Cons.} & \textbf{Bulk} & \textbf{Signif. Genes (Bulk)} \\
 & & \textbf{PCC} & \textbf{Cos. Sim.} & \textbf{PCC} & \textbf{AUROC} & \textbf{(FDR $<$ 0.05)} \\
\midrule
Proposed Model (Full)             & $1{,}724$ & $0.881 \pm 0.010$ & $0.637 \pm 0.014$ & $0.893 \pm 0.015$ & $0.661 \pm 0.003$ & $72.6\% \pm 16.0\%$ \\
Proposed Model (w/o Consistency)  & $1{,}724$ & $0.883 \pm 0.014$ & $0.586 \pm 0.047$ & $0.691 \pm 0.034$ & $0.651 \pm 0.007$ & $69.6\% \pm 14.9\%$ \\
Proposed Model (w/o Feat. Recon.) & $1{,}724$ & $0.888 \pm 0.010$ & $-$               & $0.881 \pm 0.007$ & $0.649 \pm 0.008$ & $83.4\% \pm 15.0\%$ \\
Proposed Model (w/o TCGA)         & $-$  & $-$               & $-$               & $-$               & $-$               & $-$ \\
Proposed Model (w/o TCGA cells)    & $1{,}724$ & $0.882 \pm 0.017$ & $0.541 \pm 0.061$ & $0.865 \pm 0.026$ & $0.647 \pm 0.006$ & $79.3\% \pm 12.6\%$ \\
DeepSpot2Cell           & $-$  & $-$               & $-$               & $-$               & $-$               & $-$ \\
STFlow                  & $-$  & $-$               & $-$               & $-$               & $-$               & $-$ \\
\bottomrule
\end{tabular}
}
\end{table}

\begin{table}[H]
\centering
\caption{Average 3-fold performance (panoptic and detection quality; PQ and DQ, respectively) comparison between Classpose and CellViT models trained on PanNuke\textsuperscript{*}. The best performing metrics are in bold.}
\label{tab:classpose_vs_cellvit}
\vspace{5pt}
\begin{tabular}{lcc}
\toprule
& \textbf{CellViT} & \textbf{Classpose} \\
\cmidrule(lr){2-2} \cmidrule(lr){3-3}
\textbf{Cell type} & PQ / DQ & PQ / DQ \\
\midrule
Neoplastic   & $0.581$ / \textbf{$0.71$} & \textbf{$0.600$} / \textbf{$0.71$} \\
Inflammatory & $0.417$ / $0.59$          & \textbf{$0.544$} / \textbf{$0.65$} \\
Connective   & $0.423$ / $0.53$          & \textbf{$0.441$} / \textbf{$0.55$} \\
Dead         & $0.149$ / $0.39$          & \textbf{$0.325$} / \textbf{$0.428$} \\
Epithelial   & $0.583$ / \textbf{$0.73$} & \textbf{$0.586$} / \textbf{$0.73$} \\
\bottomrule
\end{tabular}

\par\vspace{2pt}
\raggedright\scriptsize
\textsuperscript{*}Available at \url{https://huggingface.co/datasets/RationAI/PanNuke}, accessed on May 18, 2026.
\end{table}

\end{document}